\pdfoutput=1

\documentclass[11pt]{article}

\usepackage{EACL2023}

\usepackage{times}
\usepackage{latexsym}

\usepackage[T1]{fontenc}

\usepackage[utf8]{inputenc}

\usepackage{microtype}

\usepackage{inconsolata}

\usepackage{soul}[color]
\usepackage{graphicx}
\usepackage{booktabs}
\usepackage{hyphenat}
\usepackage{multirow}
\usepackage{multicol}
\usepackage{colortbl}
\usepackage{dblfloatfix} 
\usepackage{tabularx}
\usepackage{float}
\usepackage{subfig}
\usepackage{float}
\usepackage{dblfloatfix}
\usepackage{hyperref}

\definecolor{decR}{rgb}{204,0,0}
\definecolor{incG}{RGB}{0,102,0}
\definecolor{modO}{RGB}{255,179,71}
\definecolor{tablediff}{RGB}{47,79,79}

\title{Real-Time Visual Feedback to Guide Benchmark Creation: \\A Human-and-Metric-in-the-Loop Workflow}

\author{
Anjana Arunkumar$^{1}$ \\
  \texttt{aarunku5@asu.edu} \\\And
Swaroop Mishra$^{1}$ \\
  \texttt{srmishr1@asu.edu} \\\And
Bhavdeep Sachdeva$^{1}$ \\
  \texttt{bssachde@asu.edu} \\\AND
Chitta Baral$^{1}$ \\
  \texttt{cbaral@asu.edu} \\\And
Chris Bryan$^{1}$ \\
  \texttt{cbryan16@asu.edu} \\\AND
  $^1$Arizona State University  
  }

\begin{document}
\maketitle
\begin{abstract}
Recent research has shown that language models exploit `artifacts' in benchmarks to solve tasks, rather than truly learning them, leading to inflated model performance. In pursuit of creating \textit{better benchmarks}, we propose VAIDA, a novel benchmark creation paradigm for NLP, that focuses on \textit{guiding crowdworkers}, an under-explored facet of addressing benchmark idiosyncrasies. VAIDA facilitates sample correction by providing real-time visual feedback and recommendations to improve sample quality. Our approach is domain, model, task, and metric agnostic, and constitutes a paradigm shift for robust, validated, and dynamic benchmark creation via human-and-metric-in-the-loop workflows. We evaluate via expert review and a user study with NASA TLX. We find that VAIDA decreases effort, frustration, mental, and temporal demands of crowdworkers and analysts, simultaneously increasing the performance of both user groups with a \textcolor{decR}{45.8\%} decrease in the level of artifacts in created samples. As a by-product of our user study, we observe that created samples are adversarial across models, leading to decreases of \textcolor{decR}{31.3\%} (BERT), \textcolor{decR}{22.5\%} (RoBERTa), \textcolor{decR}{14.98\%} (GPT-3 fewshot) in performance.\footnote{\label{foot-1}A video description of VAIDA, generated samples, and detailed analyses are available in the Supplemental Material.}
\end{abstract}

\begin{table*}[t]
    \scriptsize
    \centering
    \resizebox{\textwidth}{!}{%
    \begin{tabular}{p{0.2\textwidth}p{0.2\textwidth}p{0.2\textwidth}p{0.4\textwidth}}
    \toprule
    \textbf{Component Name}                    & \textbf{DQI Implication}                              & \textbf{VAIDA Usage}                                               & \textbf{Artifacts Evaluated} \\
    \midrule
    1. Vocabulary                                 & Ambiguity and diversity of a dataset's language       & Does the sample contribute new words?                              &       Sample Length \cite{wallace2019trick}, New Words Introduced \cite{yaghoub2020dynamic,larson2020iterative}, Jaccard Index between n-grams \cite{larson2019outlier}                       \\
    \midrule
    2. Inter-Sample N-gram Frequency and Relation & Word/phrase repetition and similarity between samples & Does the sample contribute new combinations of words and phrases?  &         N-gram overlap \cite{wallace2019trick,yaghoub2020dynamic}, Mean-IDF \cite{stasaski2020more}                 \\
    \midrule
    3. Inter-Sample STS                           & Syntactic, semantic, and pragmatic sentence parsing  & How similar is the hypothesis to all other premises or hypotheses? &        Multi-hop reasoning \cite{wallace2019trick}, Similarity and overlap \cite{yaghoub2020dynamic}, Diversity \cite{larson2019outlier}                       \\
    \midrule
    4. Intra-Sample Word Similarity               & Word overlap and similarity within sample statements  & How similar are all words within a sample?                         &      Coreference Resolution, Multi-hop reasoning \cite{wallace2019trick}, Word Overlap \cite{larson2020iterative}
            \\
    \midrule
    5. Intra-Sample STS                           & Phrase/sentence level overlap within a sample         & How similar is the hypothesis to the premise?                      &        N-gram repetition and overlap \cite{yaghoub2020dynamic}                     \\
    \midrule
    6. N-gram Frequency per Label                 & Distribution of samples according to annotation       & Is the hypothesis too obvious for the system?                      &        Logic and Calculations \cite{wallace2019trick}, Diversity \cite{larson2019outlier}, Outliers, Entropy \cite{stasaski2020more}                     \\
    \midrule
    7. Inter-Split STS                            & Optimal similarity between train and test samples     & Is the sample too similar to an existing sample?                   &        Entity Distractors, Novel Clues \cite{wallace2019trick}, Coverage \cite{larson2019outlier}                        \\
    \bottomrule
    \end{tabular}%
    }
    \caption{Language properties considered in DQI that indicate artifact presence, their interpretation in VAIDA, and corresponding methods used in crowdsourcing pipeline evaluation; STS: semantic textual similarity.}
    \label{tab:dqi}
\end{table*}

\section{Introduction}

Researchers invest significant effort to create benchmarks in machine learning, including ImageNet~\cite{deng2009imagenet}, SQUAD~\cite{rajpurkar2016squad}, and SNLI~\cite{bowman2015large}, as well as to develop models that solve them. \textit{Can we rely on these benchmarks?} A growing body of recent research~\cite{schwartz2017effect,poliak2018hypothesis,kaushik2018much} is revealing that models exploit spurious bias/artifacts-- unintended correlations between input and output~\cite{torralba2011unbiased} (e.g. the word `not' is associated with the label `contradiction' in Natural Language Inference (NLI)~\cite{gururangan2018annotation})-- instead of the actual underlying features, to solve many popular benchmarks. Models, therefore, fail to generalize, and experience drastic performance drops when testing with out-of-distribution (OOD) data or adversarial examples  ~\cite{bras2020adversarial,Mishra2020OurEM,mccoy2019right,zhang2019paws,larson2019evaluation,sakaguchi2019winogrande,hendrycks2016baseline}. This begs the question: \textit{Shouldn't ML researchers consequently focus on  creating `better' datasets rather than developing increasingly complex models on bias-laden benchmarks?}

Deletion of samples based on bias baseline reports-- hypothesis-only baseline in NLI~\cite{dua2019drop})-- and mitigation approaches such as AFLite~\cite{sakaguchi2019winogrande} (adversarial filtering which deletes targeted data subsets), ~\cite{clark2019don,kaushik2019learning}, have the following limitations: (i) data deletion/augmentation and residual learning do not justify the original investment in data creation, and (ii) crowdworkers are not provided adequate feedback to learn what constitutes high-quality data-- and so have additional overhead due to the manual effort involved in sample creation/validation. Furthermore, \cite{parmar2022don} show that biased samples are created even when crowdworkers are provided with an initial set of annotation instructions. One potential solution to these problems is continuous, \textit{in situ} feedback about artifacts while benchmark data is being created. \textit{To our knowledge, there are no approaches that provide real-time artifact identification, feedback, and reconciliation opportunities to data creators, nor guide them on data quality}.

\textbf{Contributions: } (i) We propose \textit{VAIDA} (Visual Analytics for Interactively Discouraging Artifacts), a novel system for benchmark creation that provides continuous visual feedback to data creators in real-time. VAIDA supports artifact identification and resolution, implicitly educating \textit{crowdworkers} and \textit{analysts} on data quality. (ii) We evaluate VAIDA empirically through expert review and a user study to understand the cognitive workload it imposes. The results indicate that VAIDA decreases mental demand, temporal demand, effort, and frustration of crowdworkers (\textcolor{decR}{31.1\%}) and analysts (\textcolor{decR}{14.3\%}); it increases their performance by \textcolor{incG}{34.6\%} and \textcolor{incG}{30.8\%} respectively, and educates crowdworkers on how to create \textit{high-quality} samples. Overall, we see a \textcolor{decR}{45.8\%} decrease in the presence of artifacts in created samples. (iii) Even though our main goal is to reduce artifacts in samples, we observe that samples created in our user study are adversarial across language models with performance decreases of \textcolor{decR}{31.3\%} (BERT), \textcolor{decR}{22.5\%} (RoBERTa), and \textcolor{decR}{14.98\%} (GPT-3 fewshot).
\section{Related Work}

This work sits at the intersection of two primary areas: (1) \textit{visual analysis of data quality} (higher presence of artifacts indicates lower quality), and (2) \textit{development of a novel data collection pipeline}.\footnote{Detailed related work is in the Supplemental Material.}

\subsection{Sample Quality and Artifacts}

Data Shapley \cite{ghorbani2019data} has been proposed as a metric to quantify the value of each training datum to the predictor performance. However, the metric might not signify bias content, as the value of training datum is quantified based on predictor performance, and biases might favor the predictor. Moreover, this approach is model and task-dependent. VAIDA uses DQI (Data Quality Index), proposed by ~\cite{Mishra2020DQIAG}, to: (i) compute the overall data quality for a benchmark with $n$ data samples, and (ii) compute the impact of a new $(n+1)^{th}$ data sample.  Table~\ref{tab:dqi} broadly defines DQI components, along with their interpretation in VAIDA, and juxtaposes them against evaluation methods used in prior works on crowdsourcing pipelines, as discussed in \ref{sub:pipelinesrw}. \cite{wang2020vibe} propose a tool for measuring and mitigating artifacts in image datasets.

\subsection{Crowdsourcing Pipelines}
\label{sub:pipelinesrw}

Several pipelines have been proposed to handle various aspects of artifact presence in samples. 

\textbf{Adversarial Sample Creation: } Pipelines such as Quizbowl \cite{wallace2019trick} and Dynabench  \cite{kiela2021dynabench}, highlight portions of text from input samples during crowdsourcing, based on how important they are for model prediction; this prompts users to alter their samples, and produce samples that can fool the model being used for evaluation \cite{talmor2022commonsenseqa}. While these provide more focused feedback compared to adversarial pipelines like ANLI \cite{nie2019adversarial}, which do not provide explicit feedback on text features, adversarial sample creation is contingent on performance against a specific model (Quizbowl for instance is evaluated against IR and RNN models, and may therefore not see significant performance drops against more powerful models). Additionally, such sample creation might introduce new artifacts over time into the dataset and doesn't always correlate with high quality-- for instance, a new entity introduced to fool a model in an adversarial sample might be the result of insufficient inductive bias, though reducing the level of spurious bias.

A similar diagnostic approach is followed for unknown unknown identification-- i.e., instances for which a model makes a high-confidence prediction that is incorrect. \cite{attenberg2015beat} and \cite{vandenhof2019hybrid} propose techniques to identify UUs, in order to discover specific areas of failure in model generalization through crowdsourcing. The detection of these instances is however, model-dependent; VAIDA addresses the occurrence of such instances by comparing sample characteristics between different labels to identify (and resolve) potential artifacts and/or under-represented features in created data. 

\textbf{Promoting Sample Diversity: } Approaches focusing on improving sample diversity have been proposed, in order to promote model generalization. \cite{yaghoub2020dynamic} use a probabilistic model to generate word recommendations for crowdworker paraphrasing. \cite{larson2019outlier} propose retaining only the top k\% of paraphrase samples that are the greatest distance away from the mean sentence embedding representation of all collected data. These `outlier' samples are then used to seed the next round of paraphrasing. \cite{larson2020iterative} iteratively constrain crowdworker writing by using a taboo list of words, that prevents the repetition of over-represented words, which are also a source of spurious bias. Additionally, \cite{stasaski2020more} assess the new sample's contribution to the diversity of the entire sub-corpus. 

\textbf{Controlled Dataset Creation:}
Previous work \cite{roit2019crowdsourcing} in controlled dataset creation trains crowdworkers, and selects a subset of the best-performing crowdworkers for actual corpus creation. Each crowdworker's work is reviewed by another crowdworker, who acts as an analyst (as per our framework) of their samples. However, in real-world dataset creation, such training and selection phases might not be possible. Additionally, the absence of a metric-in-the-loop basis for feedback provided during training can potentially bias (through trainers) the created samples.

As shown in Table \ref{tab:dqi}, DQI encompasses the aspects of artifacts studied by the aforementioned works; it further quantifies the presence of many more inter and intra-sample artifacts,\footnote{See Supplemental Material for details on artifacts that DQI identifies.} and provides a one-stop solution to address artifact impact on multiple fronts. VAIDA leverages DQI to identify artifacts, and further focuses on educating crowdworkers on exactly `why' an artifact is undesirable, as well as the impact its presence will have on the overall corpus. It is also easily extensible to incorporate additional metrics such as quality control measures \cite{ustalov2021general}, enabling benchmarking evaluation in a reproducible manner. This is in contrast to the implicit feedback provided by word recommendation and/or highlighting in prior works, based on a static set of metrics-- VAIDA facilitates the systematic elimination of artifacts without the unintentional creation of new artifacts, something that has hitherto remained unaddressed.

\section{Modules}

In this section, we describe VAIDA's important backend processes.

\textbf{DQI:} DQI can be expressed as a quality metric that examines different sources of artifacts in text, by scoring samples along 7 different components. We use DQI in order to demonstrate VAIDA's ability to cover multiple facets of artifact creation, although VAIDA is metric agnostic. VAIDA uses an intuitive traffic signal color coding (\textcolor{decR}{high}$>>$\textcolor{modO}{moderate}$>>$\textcolor{incG}{low}) to indicate levels of artifacts (i.e., quality) in samples. Hyperparameters for color mapping with respect to DQI component values depend on (i) the application type, and (ii) characteristics of pre-existing samples present in the corpus at the time of new sample creation.\footnote{\label{footeval} \label{footref} See Supplemental Material: Evaluation, for details  across all DQI components, hyperparameter tuning, and analyses.} For instance, when recreating SNLI \cite{bowman2015large} with VAIDA, we tune hyperparameters separating the boundary between red, yellow, and green flags on 0.01\% of the SNLI training dataset manually in a supervised manner \cite{Mishra2020DQIAG}.

\definecolor{myRed}{rgb}{0.97,0.36,0.3}
\definecolor{myYellow}{rgb}{0.96,0.74,0.37}
\definecolor{myGreen}{rgb}{0.56,0.93,0.56}

\begin{figure}[h]
\centering
\includegraphics[width=\columnwidth]{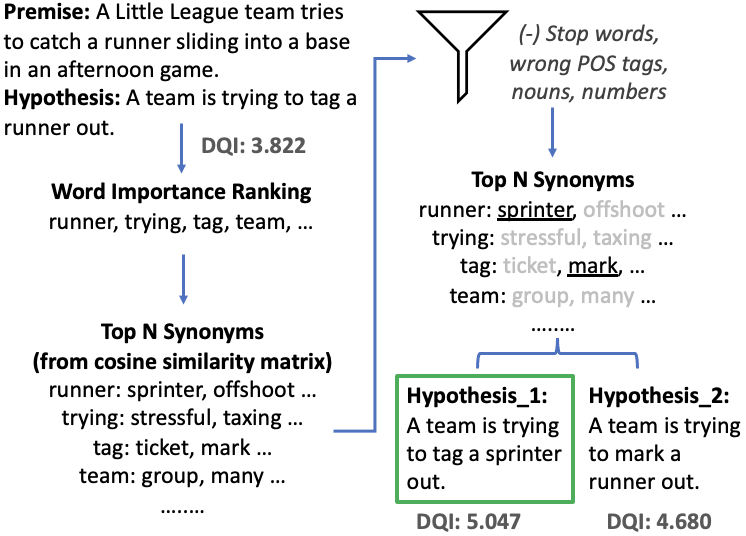}
  \caption{AutoFix Algorithm applied to an SNLI entailment sample, replacing one word per iteration. The DQI of the hypothesis changes from 3.822 to 5.047.}
\label{fig:AutoFix}
\end{figure}

\label{sec:auto}
\textbf{AutoFix: } We propose AutoFix as a module to help crowdworkers avoid creating bad samples by recommending changes to a sample to improve its quality. The AutoFix algorithm is explained in Figure~\ref{fig:AutoFix}. Given a premise, hypothesis, and DQI values for the hypothesis, AutoFix sequentially masks each word in the hypothesis and ranks words based on their influence on model output, i.e., their importance. The hypothesis word of highest importance is replaced, to achieve at least \textcolor{modO}{moderate} quality. DQI hence controls the amount and aspect of changes made by AutoFix. By incrementally changing the sample, users can understand how and why their sample is being modified and how DQI values are affected.

\label{sec:tf}
\textbf{TextFooler: } From an analyst's perspective, the quality of a submitted sample might be ``too low'' because (i) the crowdworker might not employ AutoFix appropriately, or (ii) there is a narrow acceptability range due to the criticality of the application domain, such as in BioNLP~\cite{lee2020biobert, parmar2022boxbart}. We therefore implement TextFooler~\cite{jin2019bert} to adversarially transform low-quality samples (instead of discarding them), to improve benchmark robustness, and ensure that crowdsourcing effort is not wasted. We initially use AFLite~\cite{bras2020adversarial}, to bin samples into \textit{good} (retained samples) and \textit{bad} (filtered samples) splits. Using TextFooler, we adversarially transform bad split data to flip the label; we revert back to the original label and identify sample artifacts using DQI (see Figure \ref{fig:wf2}).

\section{Interface Design and Workflow}
\label{interface}

\begin{figure*}[!h]
    \centering
    \includegraphics[width=1.67\columnwidth]{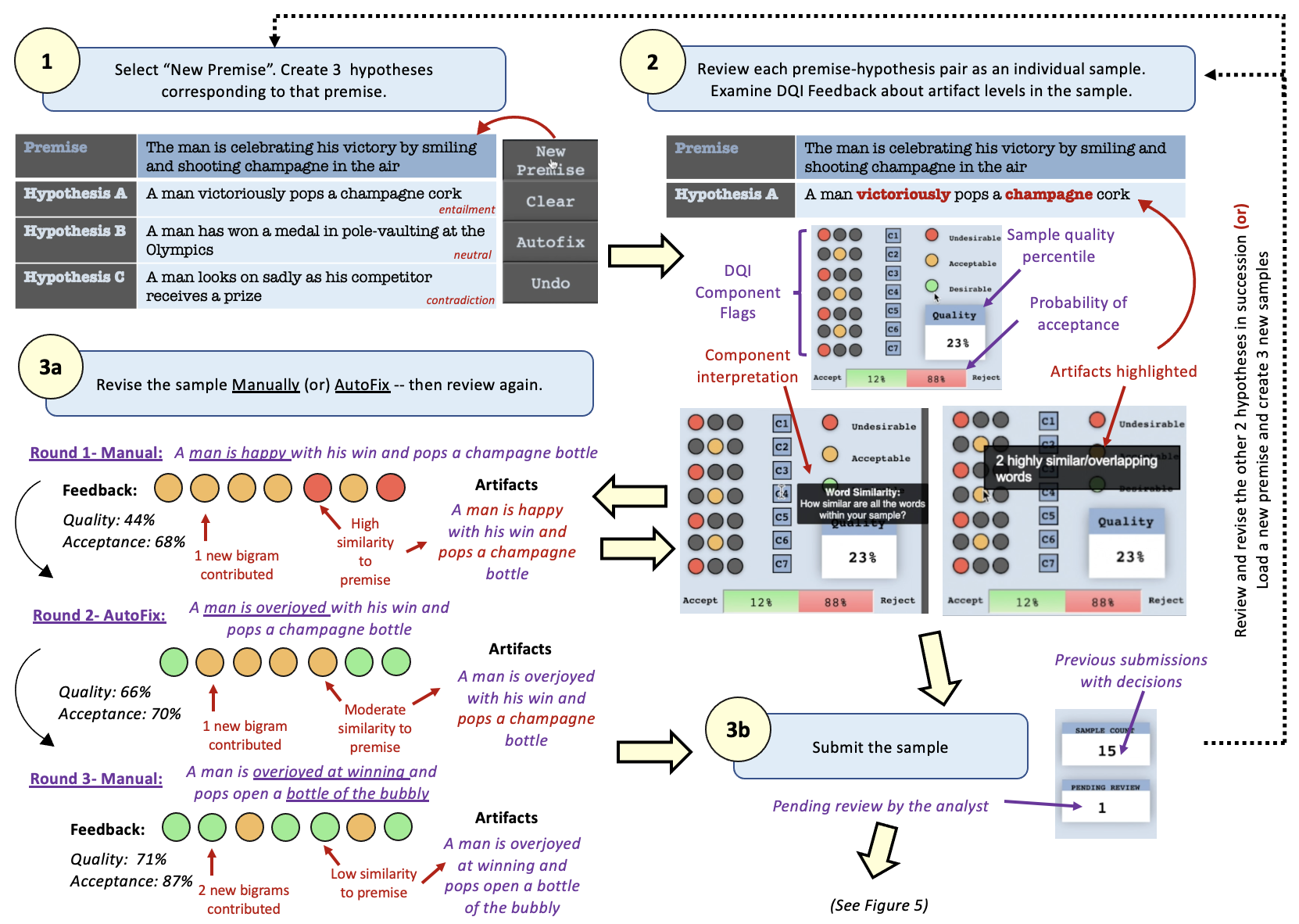}
    \caption{VAIDA workflow for the creation of a single sample by a crowdworker.}
    \label{fig:wf1}
\end{figure*}

\begin{figure*}[!h]
    \centering
    \subfloat{\includegraphics[width=1.67\columnwidth]{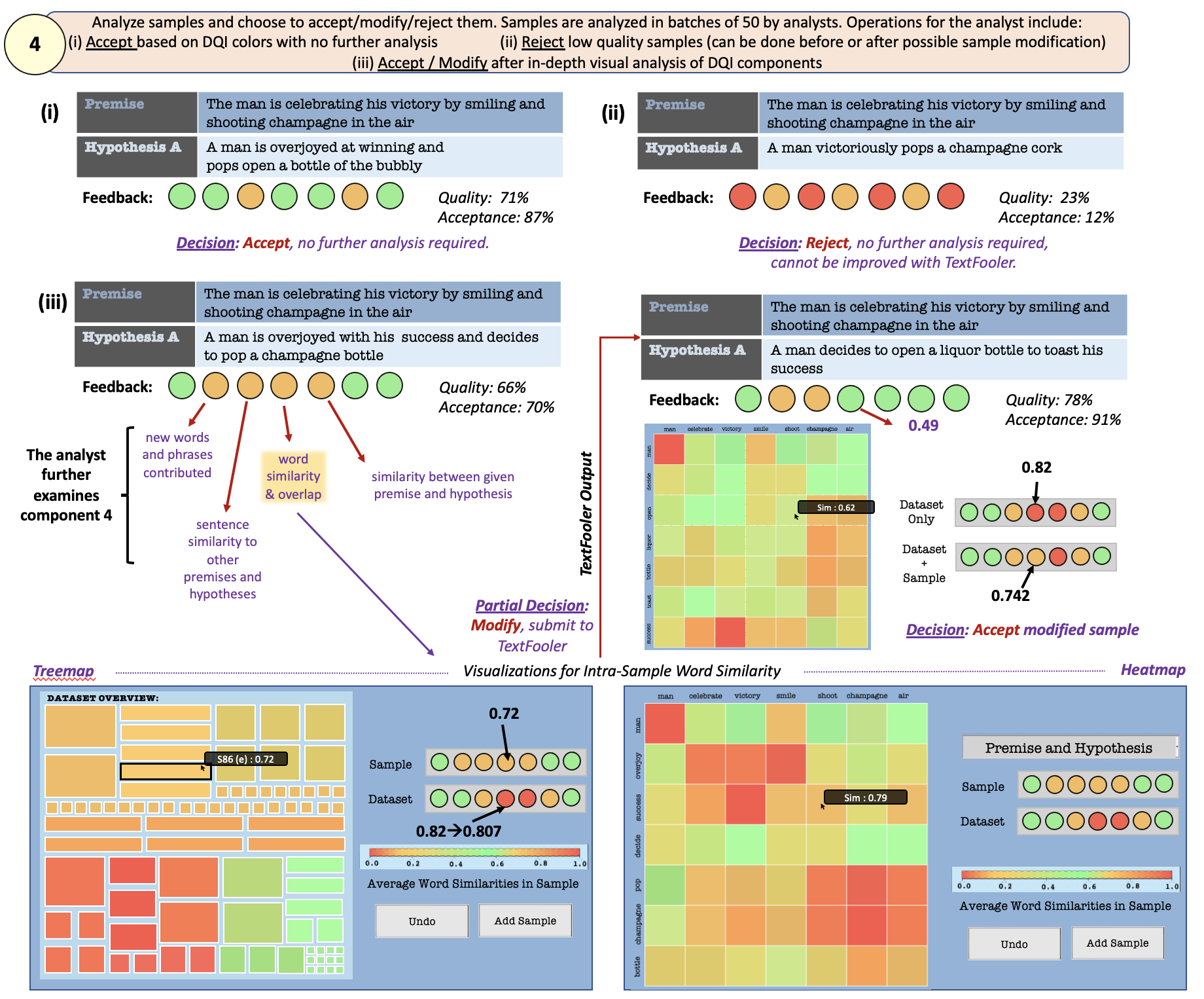}}\\
    \subfloat{\includegraphics[width=1.67\columnwidth]{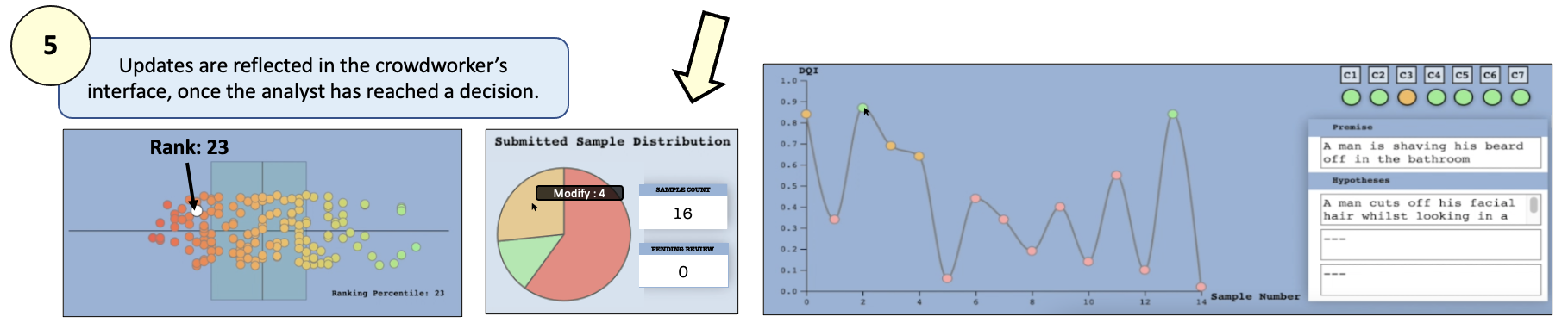}}
    \caption{VAIDA workflow for the evaluation of a single sample by an analyst, and subsequent analyst feedback provided to the crowdworker.}
    \label{fig:wf2}
\end{figure*}

VAIDA provides customized interfaces for both crowdworkers and analysts, as shown in Figures~\ref{fig:crowdworker_interface}, ~\ref{fig:analyst_interfaces} respectively.\footnote{\label{footi}See Supplemental Material: Interface Design for interface intuitions and detailed descriptions, with full-resolution images.} We describe VAIDA's workflow via a case study for sample creation (crowdworker) and review (analyst) in Figures~\ref{fig:wf1},\ref{fig:wf2}.

 \begin{figure}[h]
    \centering
    \includegraphics[width=.5\textwidth]{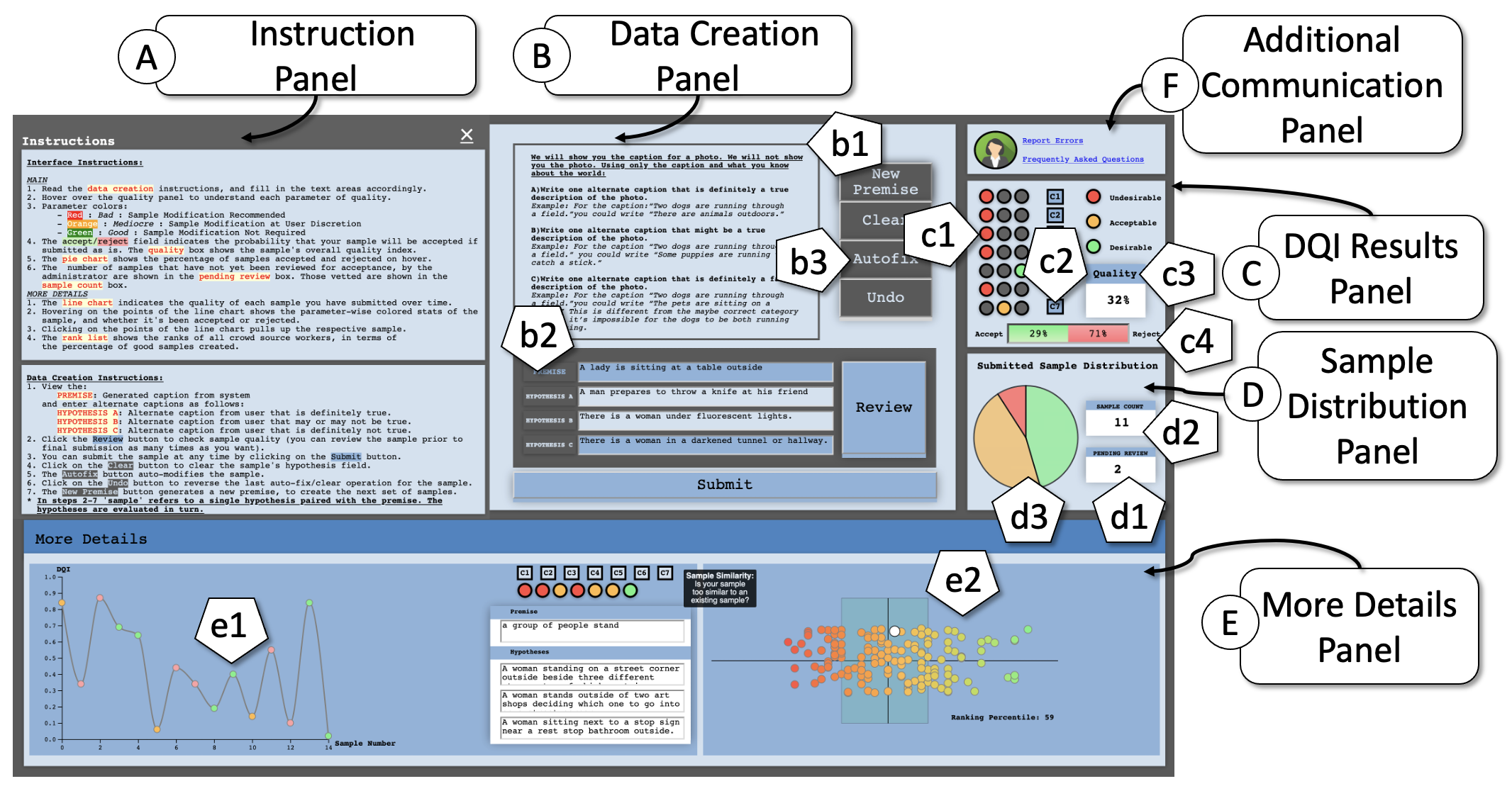}
    \caption{VAIDA's crowdworker interface consists of six linked panels: (A) Instructions, (B) Data creation, (C) DQI results, (D) Sample distribution, (E) More details, and (F) Additional communication.\textsuperscript{\ref{footi}}}
    \label{fig:crowdworker_interface}
\end{figure}

 \begin{figure}[h]
    \centering
    \includegraphics[width=.5\textwidth]{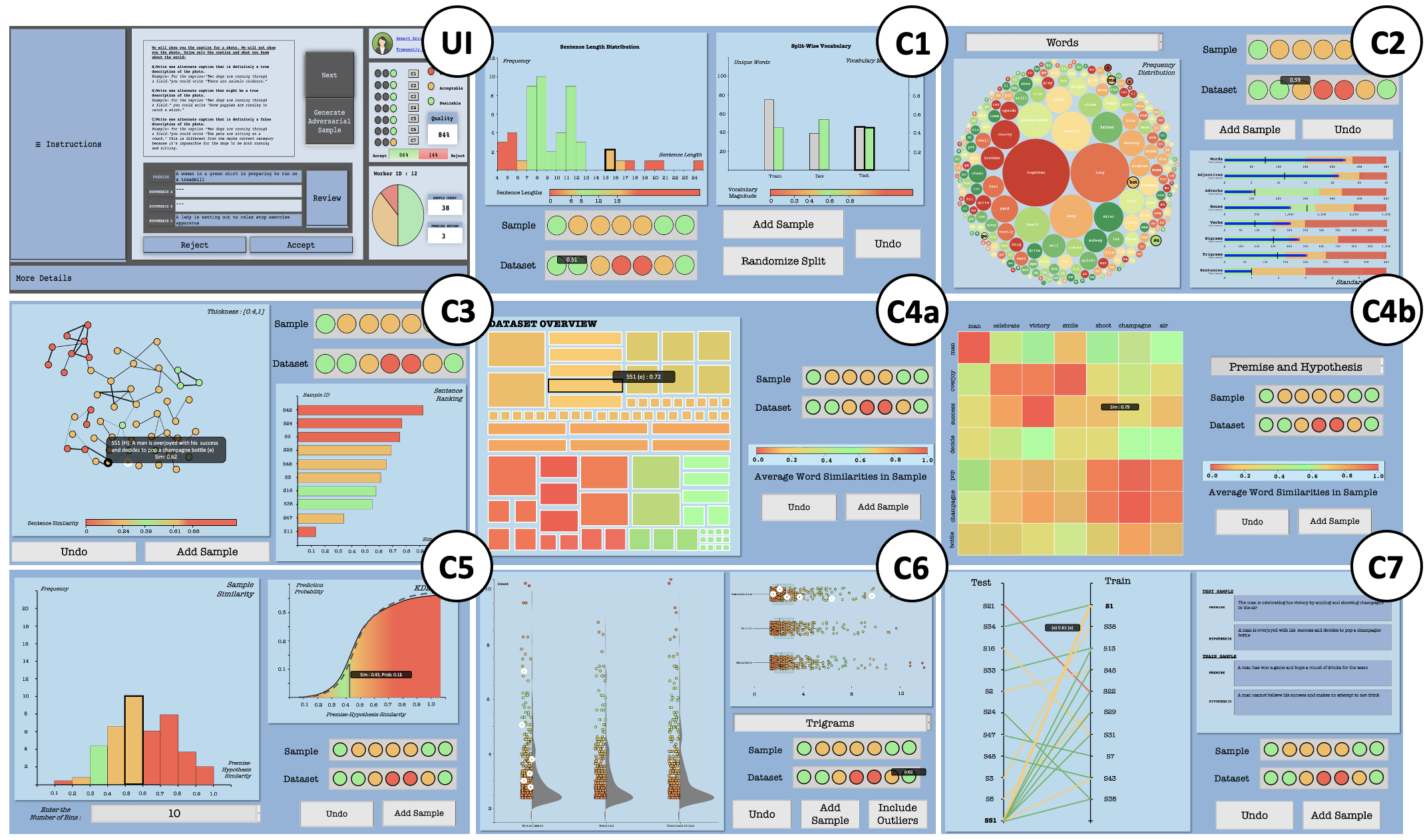}
    \caption{VAIDA provides a collection of interfaces for the analyst supporting detailed analysis and investigation of submitted samples and the overall benchmark.\textsuperscript{\ref{footi}}}
    \label{fig:analyst_interfaces}
\end{figure}

The crowdworker interface provides instructions \textbf{(A)} to navigate through the panels and how to interpret feedback for created samples. Communication links for FAQs, and error reporting are also provided (F). Crowdworkers can then review the instructions for data creation \textbf{(B)}-- here, we use the same instructions provided in the original SNLI crowdsourcing interface \textbf{(b1)}. 

\textbf{1. Sample Creation:} The premise field auto-populates with a caption from the Flickr30 corpus \textbf{(b2)}. The crowdworker must create three hypotheses (for the entailment, neutral, contradiction labels) at a time, though they are reviewed individually.

\textbf{2. DQI Evaluation Feedback:} On clicking the review button, DQI feedback is shown in \textbf{(C)}, for each component. Each colored circle \textbf{(c1)} indicates the level of artifacts present corresponding to each DQI component for the created sample; hovering displays a tooltip that explains and highlights \textbf{(c2)} the artifacts in the created sample, pertaining to the category of bias covered by that component. The overall sample quality \textbf{(c3)} is calculated, by averaging over the artifact percentiles of all 7 DQI components for the sample. By comparing sample quality with that of pre-existing samples, the probability that the sample will be accepted/rejected is shown \textbf{(c4)}. The user can choose to either revise or submit the sample.

\textbf{3a. Sample Revision:} \textit{Manual Revision} or \textit{AutoFix} can be done. We illustrate the improvement of sample quality in this step in Figure~\ref{fig:wf1}.

\textbf{3b. Sample Submission:} After review (and potentially iterative DQI evaluations/sample revisions), the sample can be submitted for benchmark inclusion. On submission, a sample enters a \textit{pending} state \textbf{(d1)} until review by the analyst.

While crowdworkers work within a single tightly-coordinated interface to create, submit, and review samples, analysts can navigate between a set of nine screens (Figure~\ref{fig:analyst_interfaces}) to review samples in detail to make `accept', `reject', and `modification' decisions, and to assess overall benchmark quality. Analysts review samples in batches of size 50.\footnote{DQI components 1, 2, 3, 6, and 7, gauge artifact presence relative to pre-existing samples. We observe given at least 50 pre-existing samples, DQI component values change by less than 5\% for the overall dataset if another 50 samples are accepted.}

\textbf{4. Analyst Review:} Review can consist of several different operations by the analyst.

\textit{(i) Direct Acceptance--} The home page \textbf{(UI)} for the analyst provides a view similar to the crowdworker interface, and allows the analyst to review the work of a single crowdworker. If the data quality is deemed to be sufficiently high by just viewing the DQI color flags and quality percentile, the analyst can directly accept the sample into the corpus from this screen, as shown in Figure~\ref{fig:wf2}.

\textit{(ii) Visual Analysis and Modification--} VAIDA provides several visualizations \textbf{(C1-C7)} to support detailed analysis and review of submitted samples, and to assess artifact presence (i.e., quality) in the overall benchmark. Each visualization allows the analyst to simulate how adding one or more submitted samples affects the benchmark's quality.

For example, in Figure~\ref{fig:wf2}, we show the visualizations for analysis of DQI component 4 (C4), which deals with artifacts caused due to word similarity within a sample. For each sample in the corpus, the word similarities between all possible pairs of words are averaged; the samples are then hierarchically displayed as a treemap based on their DQI color mapping and average word similarity. We note that the color scale followed is bilinear, as while artifacts must be eliminated, there still needs to be a sufficient inductive bias for the sample to be solvable. The size of a rectangle indicates the distance of each sample from the average word similarity across all samples. The new sample is highlighted in the treemap with a black outline. In the example shown, we see that the dataset's C4 value decreases slightly (0.807) when the new sample is added, though the flag color remains red.

Further examination of the new sample can be done, to establish why the sample has low effect on the dataset's C4; the user clicks on the sample rectangle in the treemap, and is taken to a heat map view. The heatmap shows the similarity values between every word pair (from the premise/hypothesis/both) in a tooltip on hover, with the values mapped to the same color scale used for the treemap . For instance, in Figure~\ref{fig:wf2}, the words `man' and `champagne are repeated verbatim, and `overjoyed', `smiling', `celebrating', `pop', and `shoot' also have similarity of [0.6--1]. This indicates that several words in the sample are too closely related and constitute artifacts. 

Each such visualization is therefore individually tailored to represent a specific DQI component of interest, based on the linguistic features examined for artifact creation in that component. We further elaborate on the design intuitions and analysis conducted with all the component visualizations available to the analyst in the supplementary.

Post analysis, if the analyst feels that the submitted sample requires only a minor change (for instance, reshuffling or the addition of a single word) to warrant acceptance, then they can invoke the TextFooler module to transform the sample adversarially, and then accept, thereby ensuring minimal data loss. In the case shown, TextFooler improves the sample's C4 with most of the heatmap varying from 0.3-0.7. Due to this, the analyst decides to accept the updated sample.

\textbf{5. Analyst Feedback:} Once the analyst has reached a decision, the crowdworker sees updates (Figure \ref{fig:wf2}) in (D) to the reviewed sample count (d2) -- increases to 16-- and the pie chart that indicates the distribution of actions taken by the analyst over all samples submitted by the crowdworker (d3). Additionally, in (E), the line chart (e1), which contains the history of previously submitted samples is updated. The x-axis denotes the sample number and the y-axis denotes the quality percentile (c3), of the corresponding sample. On click, the corresponding sample is loaded, along with its DQI component values and feedback to the crowdworker. The crowdworker can also use the box plot (e2), to view their current rank, and choose to view the sample history of another crowd worker.
\section{Evaluation}

We evaluate VAIDA's efficacy at providing real-time feedback to educate crowdworkers during benchmark creation using expert review and a user study. We also evaluate model performance (BERT~\cite{devlin2018bert}, RoBERTa~\cite{liu2019roberta}, GPT-3 (fewshot)~\cite{brown2020language}) on data created with VAIDA during the user study.

\subsection{Expert Review}
We present an initial prototype of our tool, to a set of three researchers  with expertise in NLP and knowledge of data visualization. For each expert, the two interfaces were demoed in a Pair-Analytics session \cite{5718616}. Participants could ask questions and make interaction/navigation decisions to facilitate a natural user experience. All the experts appreciated the easily interpretable traffic-signal color scheme (and further suggested that alternates be provided to account for color blindness) and found the organization of the interfaces---providing separate detailed views within the analyst workflow--  a way to prevent cognitive overload (too much information on one screen). A caveat of this would be the inability for an analyst to simultaneously juxtapose different component visualizations. It was also hypothesized that a learning curve of ~$\sim$50--60 samples would be required for cohesive use of all system modules by both types of users; however, this would be offset by the eventual capability of users to deal with samples of middling quality based on their multi-granular feedback about artifact presence.

\subsection{User Study}\label{sec-4.3}
\begin{figure}[h]
    \centering
    \includegraphics[width=0.9\columnwidth]{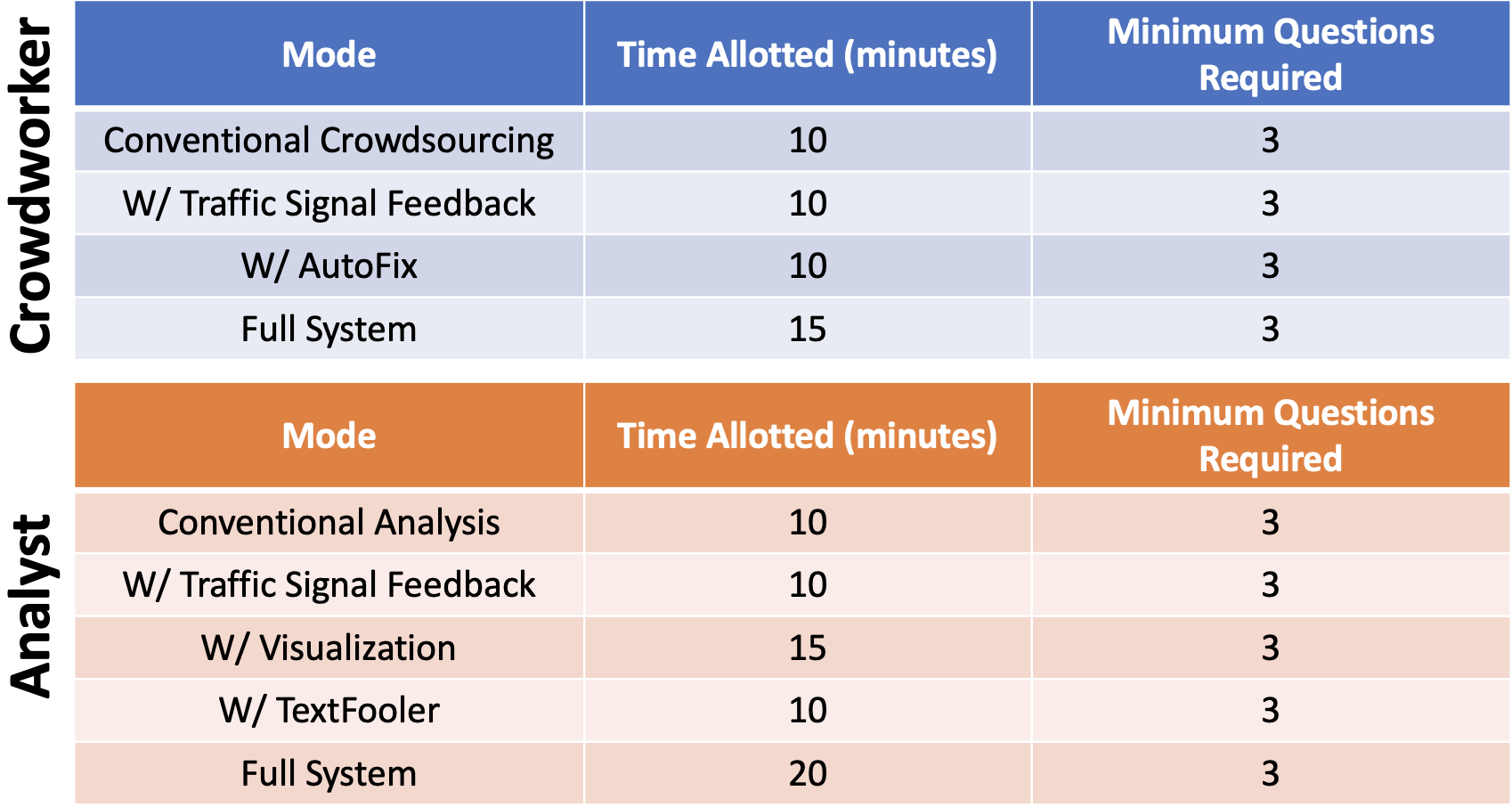}
    \caption{User Study Setup-- describes the timeframe and requirements of the user study over ablation rounds.}
    \label{user1}
\end{figure}

\textbf{Setup: } We approach several software developers, testing managers, and undergraduate/graduate students. Based on their domain familiarity (in NLP and visualization, rated from 1:novice-5:expert), we split them into 23 crowdworkers and 8 analysts for constructing NLI samples, given premises. There are 100 high-quality samples in the system at the time each participant participates in each ablation round (Table \ref{tab:conf}). For both types of users, a preliminary walkthrough of the system configuration, using 2 fixed samples, is conducted for each round of the study (Figure \ref{user1}). At the end of each round, they are also asked for their comments.\textsuperscript{\ref{foot1}}

\begin{table}[h]
    \centering
    \scriptsize
    \resizebox{\columnwidth}{!}{%
    \begin{tabular}{p{0.2\columnwidth}p{0.725\columnwidth}p{0.075\columnwidth}}
        \toprule
         \textbf{Configuration} & \textbf{Description} & \textbf{User}\\\midrule
         Conventional Crowdsourcing & No feedback or auto modification tools& C \\\midrule
         Conventional Analysis & Manual review without feedback or modification tools & A \\\midrule
         Traffic Signal Feedback & Color mapping based on DQI values & C, A \\\midrule
         AutoFix & Incremental sample auto-modification functionality& C \\\midrule
         TextFooler & Adversarial sample transformation functionality & A \\\midrule
         Visualization & Data visualizations for in-depth exploration (also includes traffic signal feedback) & A \\\midrule
         Full System & All modules and system functionalities & C,A \\\bottomrule
    \end{tabular}%
    }
    \caption{System configurations used for randomly ordered ablation rounds presented to users functioning as crowdworkers (C) and analysts (A) in the study.}
    \label{tab:conf}
\end{table}

User experience (mental workload) is subjectively evaluated using NASA Task Load Index~\cite{hart2006nasa}\footnote{ \label{foot1} See Supplemental Material: User Study for more details. We do aggregated analysis of comments, full quotes of comments are present in the Supplemental Material. We also have IRB approval to conduct this user study.} (NASA TLX); each dimension is scored in a 100-points range, with 5-point steps. Users are also asked to report overall ratings for each system configuration at the end of the study.

\begin{figure}[t]
    \centering
    \includegraphics[width=0.85\columnwidth]{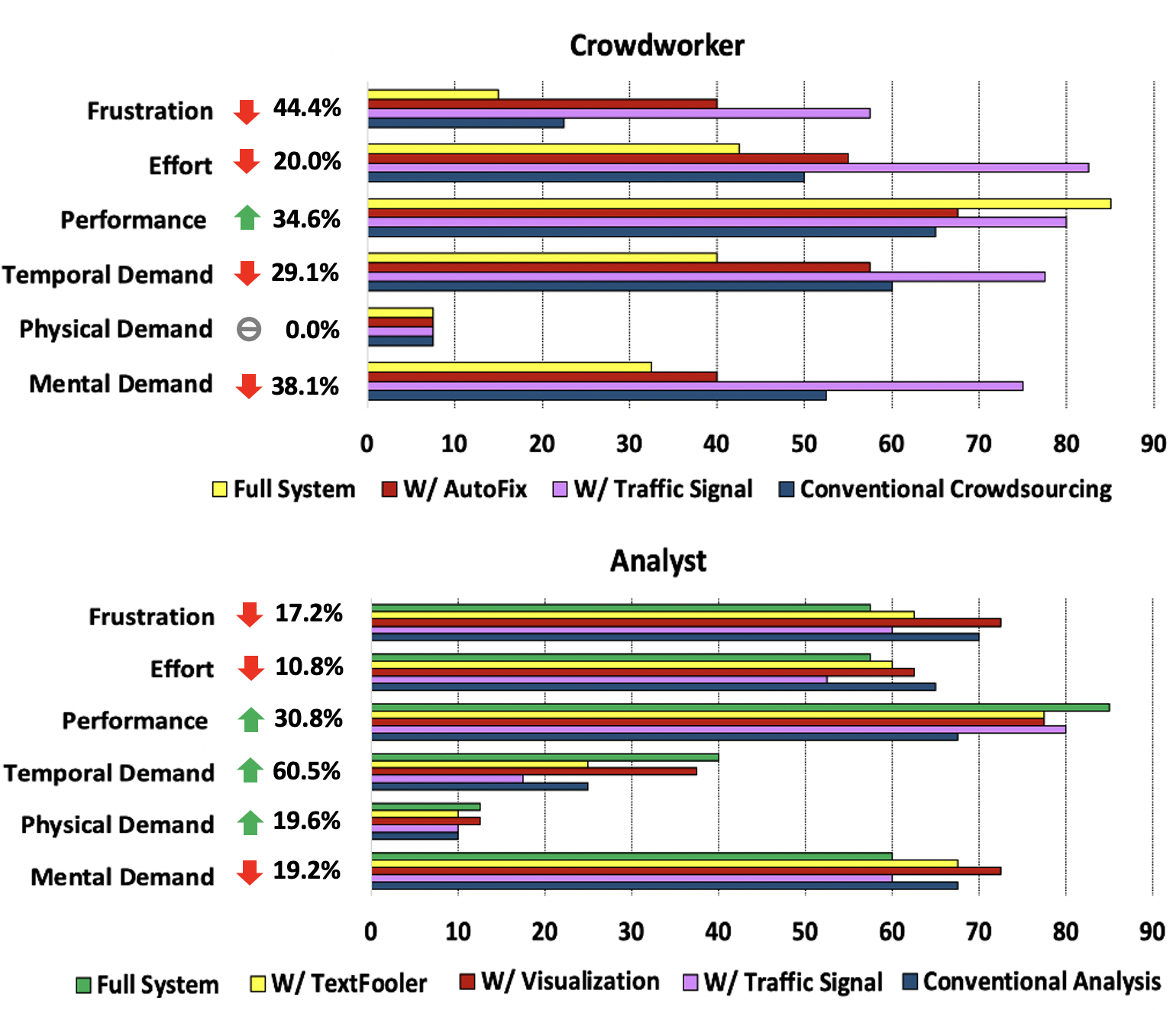}
    \includegraphics[width=0.85\columnwidth]{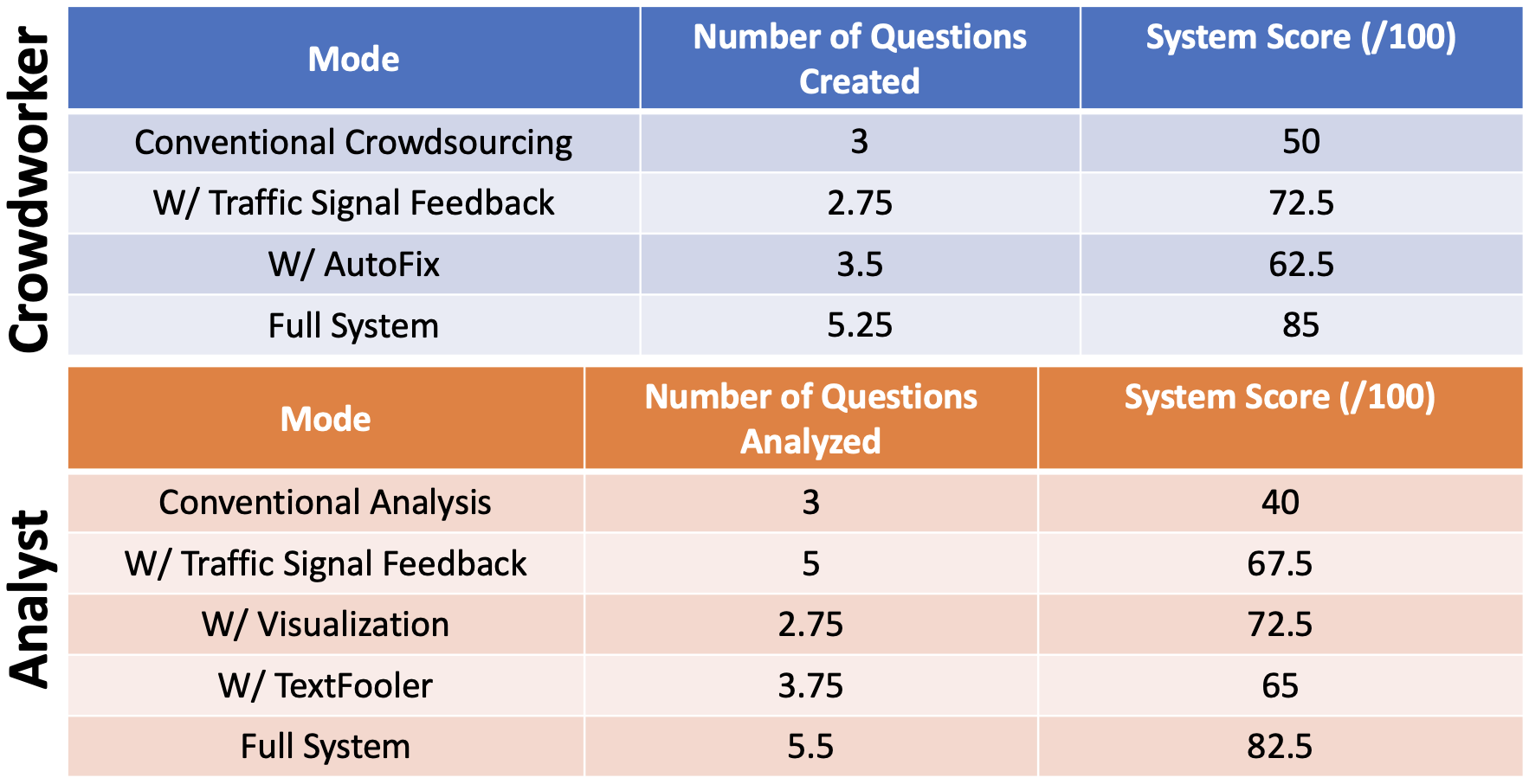}
    \caption{User Study Results-- NASA TLX dimension scores, and user scoring of the configurations, averaged across participant responses, for each ablation round. Percentages shown indicate comparison of full system scores with respect to the conventional system.}
    \label{user2}
\end{figure}

\textbf{Analysis: } Figure \ref{user2} summarizes study results,\footnote{All results are found to have $p \leq 0.02$} averaged over all user responses, for different system configurations. The general trend across both crowdworkers and analysts is that there is: (i) significant improvement across all NASA TLX dimensions, (ii) increase in number of samples created/reviewed, and (iii) user ratings for the system, when comparing VAIDA to conventional interfaces. In the case of partial module availability, we find that the effectiveness of traffic signal feedback and visualizations is comparable. The use of AutoFix and TextFooler\ref{foot1} is more prevalent initially, on creation/evaluation of a low or middling quality sample for users as: (i) crowdworkers find constructive sample modification more difficult initially, and (ii) analysts are initially unsure of how to deal with middling quality samples.

\begin{figure*}[t]
    \centering
    \includegraphics[width=0.82\textwidth]{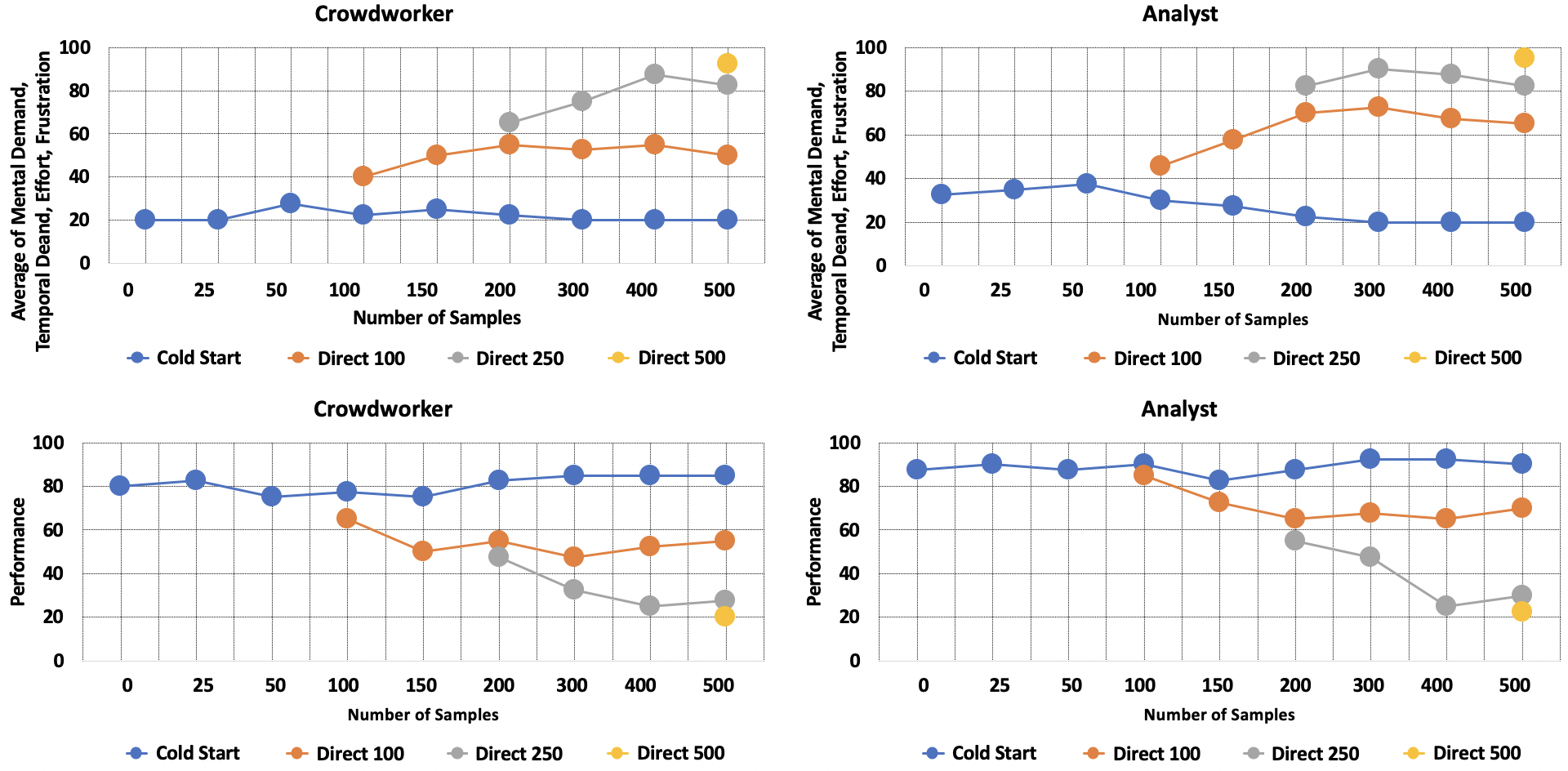}
    \caption{User education curves. Cold start has no pre-existing samples, and direct-n has n pre-existing samples. Mental Demand, Temporal Demand, Frustration, and Effort are averaged, Physical Demand is ignored. Performance is plotted separately as it shows differing behavior from the others.}
    \label{figuser}
\end{figure*}

\textbf{Learning Curve: } At the end of the study, all users are asked the following: \textit{``What do you think high quality means?"} We find that users are able to distinguish  certain patterns that promote higher quality, such as keeping sentence length appropriate and uniform across labels (not too long/short), using complex phrasing (`not bad')/gender information/modifiers across labels, decreasing premise-hypothesis word overlap; they also do not display undesirable behavior like tweaking previously submitted samples just to create more. We also find an overall decrease of \textcolor{decR}{-45.8\%} in the level of artifacts of created samples, across all rounds of ablation.

\textbf{User Education: }We also conduct a secondary study where a subset of participants (7 crowdworkers and 2 analysts) agreed to create/ analyze samples, for varying numbers of pre-accepted samples (Figure \ref{figuser}), in only the full system condition. We find that when participants are directly started in situations with $>$ 500 samples in the system, their unfamiliarity with the system initially causes a steepening of the learning curve compared to the cold start condition; this also tapers and saturates more slowly than cold start as the users gain experience. This is attributed to: (a) an increased likelihood of samples of low/middling quality (more artifacts) being created (evinced by performance), and (b) lower impact of an individual sample on overall dataset quality. We also find that users who create $\sim$50 samples report lesser reliance on AutoFix as they get better at creating higher quality samples; those who analyze $\sim$75 samples use TextFooler more efficiently as they understand how to deal with samples of middling quality better in the cold start condition. These numbers increase by $\sim$25\% when users start with 500 pre-existing samples. 

\begin{figure}[h]
    \centering
    \includegraphics[width=\columnwidth]{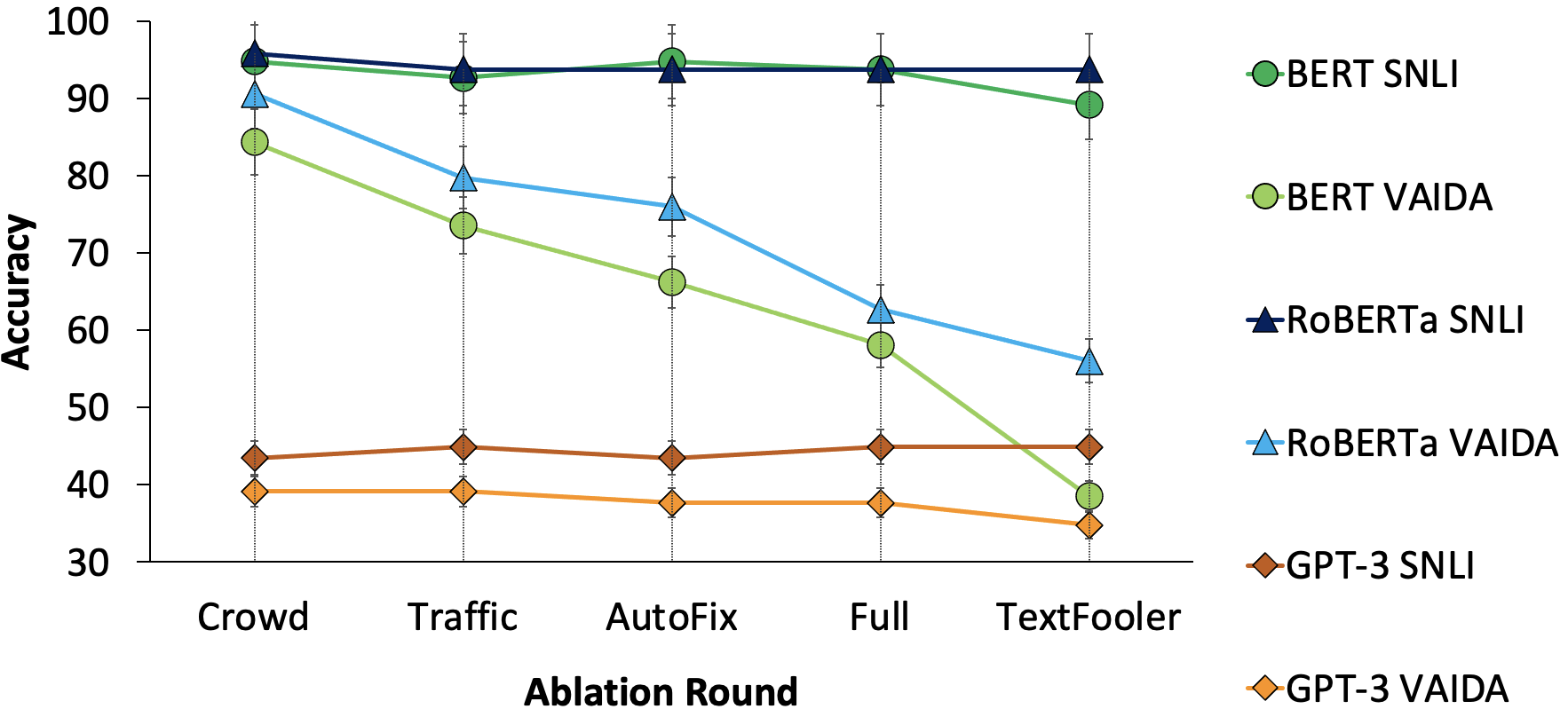}
    \caption{Model performance results for samples created during each ablation round of the user study.}
    \label{fig:model}
\end{figure}

\subsection{Model Performance Results}
We evaluate BERT and RoBERTa (trained on the full SNLI dataset), and GPT-3 (in fewshot setting) against the data created during the ablation rounds of the user study.\footnote{\label{scd}We also create samples for the Story CLOZE dataset in the full system condition; see Supplementary Material} There are 100 high-quality samples(DQI>0.7) from the original SNLI dataset present in the system for the study\textsuperscript{\ref{mprd}}. This remains constant across all ablation rounds and for all users. 69 samples are created per ablation round during our study, across all users, for a total of 345(69*5) samples that we evaluate with the models. Figure \ref{fig:model} shows the results for samples over each round of ablation. (\footnote{\label{mprd}The dataset is included in the Supplemental Material.}) In the case of TextFooler, samples are created using the `full system' condition and then further modified using TextFooler by the analyst. The other sample sets are not modified by the analyst, and are directly accepted after evaluation. 

We find that across all models, performance is lower when explicit quality feedback (via the traffic signal scheme) is provided, compared to the regular crowdsourcing condition. The highest drop is seen for BERT (\textcolor{decR}{-20.66\%}), while GPT-3 shows a lesser magnitude of performance decrease (\textcolor{decR}{-12.89\%}). Performance further decreases for all models when AutoFix is implemented, indicating the effectiveness of this module in seeding suggestions for sample improvement; the magnitude of performance loss follows BERT$>>$RoBERTa$>>$GPT-3. We can attribute this apparent variation in model robustness i.e., BERT$<<$RoBERTa$<<$GPT-3 as proportionate to increase in size of (i) the respective language models, and (ii) the pre-training corpora.

A significant decrease is seen in the full system and TextFooler conditions. Particularly, in the TextFooler round, performance sharply decreases for all models (\textcolor{decR}{-31.3\%} BERT, \textcolor{decR}{-22.5\%} RoBERTa, \textcolor{decR}{-14.98\%} GPT-3). Furthermore, in the TextFooler round, there is a \textcolor{decR}{-71.70\%} decrease in the level of sample artifacts. This indicates that crowdworkers and analysts are able to utilize VAIDA's affordances to create more robust text samples.
\section{Discussion and Conclusion}

We propose VAIDA, a paradigm to address benchmark artifacts, by integrating human-in-the-loop sensemaking with continuous visual feedback. VAIDA uses several visualization interfaces to analyze quality considerations (based on artifact levels) at multiple granularities. While we do not explicitly address computational quality control or fairness consideration (though some aspects can be targeted by currently integrated metrics), since VAIDA is extensible to the incorporation of customized backend-metrics, we believe our paradigm can support multi-faceted benchmark evaluation. 

In our usability evaluation, we see that users report greater satisfaction, and lower difficulty with their work and system experience; this implies possible higher crowdworker retention and engagement. Additionally, in our study, we see that users effectively identify and avoid artifact patterns during sample creation. Based on our study results, we believe that a minimum of 30 annotators would be needed for large-scale data creation to ensure timely feedback (i.e., sample decision provided within 24 hours) to crowdworkers. Based on our secondary study, we believe analysts will exhibit increased performance and maintain satisfaction ratings in full-scale creation, as they will become well-versed in the nuances of bias for the data-creation task they are evaluating, as well as the visualizations being used. This, however, is contingent on restricting the visualization views to display only the ~200--300 samples with closest artifact levels to the sample being evaluated.

Overall, samples created with VAIDA are found to not only of higher quality than achieved with conventional crowdsourcing, but are also seen to be adversarial across transformer models. This is also maintained across multiple task types-- we additionally create StoryCLOZE \cite{schwartz2017effect} samples with VAIDA\textsuperscript{\ref{scd}}. This was done independent of the study described in the paper, with 4 crowdworkers creating samples. However, for this, several interface features had to be changed, so we focus on reporting only NLI results in the main paper. VAIDA hence demonstrates a novel, dynamic approach for building benchmarks and mitigating artifacts, and serves as a starting point for the next generation of benchmarks in machine learning. 

\newpage

\section*{Limitations}

In future work, we intend to integrate VAIDA with an actual crowdsourcing framework, and run a full-scale data creation study to create a high-quality benchmark. Expanding to such a setup will require additional back-end engineering, to ensure that (i) timely and accurate feedback continues to be provided in real-time to crowdworkers, (ii) analysts are available on hand to process samples in a timely manner.  This is out of scope for the current paper (e.g., it would require a significant budget), but we see our current work as a stepping stone in this direction. Additionally, studying the problem and designing the visualizations and real-time feedback mechanisms are essential steps before moving to large-scale evaluation; the novel affordances and designs are a necessary first step, and we believe they will be impactful to the NLP community. 

Crowdworker retention and engagement in this full-scale setting also need to be evaluated, in order to better contextualize the learning curve associated with system usage and handling artifact creation, given an increasingly higher number of pre-existing system samples. Comparing this setup directly with the effect of in-depth user training \cite{roit2019crowdsourcing} on artifact creation and review, prior to crowdsourcing, would also further help analyze and quantify if/how user strategy and performance changes during VAIDA usage. 

Design modifications when creating different types of datasets will mainly require the redesigning of sample input fields, corresponding to the application and the type of metric used for artifact evaluation. However, in full-scale dataset creation, the visualization views for the analyst corresponding to different artifact types will have to be restricted to the $\sim$300 samples closest artifact levels, to the given sample being created, in order to facilitate scalable processing for analysts. Additionally, since visualization familiarity is required for the analyst to effectively review samples, the analysts may choose to streamline analysis by only using a subset of the provided visualization types in their version of the system, corresponding to the application domain.

\bibliography{anthology,custom}
\bibliographystyle{acl_natbib}
\appendix

\section{Supplementary Material}
\label{sec:appendix}

The following information is included in the appendix.

\begin{itemize}
    \item \hyperref[supp1]{Infrastructure Used}
    \item \hyperref[supp2]{Run-time Estimations}
    \item \hyperref[supp3]{Hyper Parameter}
    \item \hyperref[supp4]{Related Work}
    \item \hyperref[supp5]{DQI Components}
    \item \hyperref[supp7]{Interface Design Intuitions}
    \item \hyperref[supp8]{AutoFix and TextFooler Examples}
    \item \hyperref[supp9]{User Study}
    \item \hyperref[supp10]{Expert and User Comments}
\end{itemize}

Please refer to \url{https://github.com/aarunku5/VAIDA-EACL-2023.git} for:

\begin{itemize}
    \item {Video demos of VAIDA workflow}
    \item {Sample dataset generated during the ablation rounds of the user study}
    \item {DQI and Model Performance Results for User Study Samples}
    \item {DQI Evaluation: Artifact Case Study}
\end{itemize}

\subsection{Infrastructure Used }\label{supp1}
In Section 3, we describe VAIDA's flow by high level workflow and back-end processes(DQI, AutoFix, and TextFooler). Further, as discussed in Subsection 3.2 DQI can be used for quantifying artifact presence for the: i) overall benchmark, and ii) impact of new samples. Depending on the task at hand we run our experiments in different hardware settings. The DQI calculations run mostly using CPU, for new samples as well as overall samples. The AutoFix procedure, as explained in Subsection 3.2, gives the user assistance in improving quality on a per submission basis. Therefore that does not require high GPU intensive systems; we have provisions to shift execution to a GPU as well if necessary to speed up the process. For TextFooler the fine tuning of the model is run on "TeslaV100-SXM2-16GB"; CPU cores per node 20; CPU memory per node: 95,142 MB; CPU memory per core: 4,757 MB-- this is not a necessity as code has been tested on lower configuration GPUs as well but we have run our experiments in this setting. The attack part of the TextFooler requires more memory and we run that code on "Tesla V100-SXM2-32GB" com-pute Capability:  7.0 core Clock:  1.53GHz, coreCount:  80,  device Memory Size:  31.75GiB device Memory Bandwidth:  836.37GiB/s.

\subsection{Run-time Estimations}\label{supp2}
The DQI calculations run on CPU (for real life setting purposes); for the approximate estimate of the time taken, we run experiments for fixed data size of 10K samples. If the DQI calculations are done to calculate the impact of individual new samples it take a couple of seconds. On the other hand, If we take the whole 10k size dataset it takes around 48 hours to complete the process on CPU. This whole process can be run in parallel to reduce the time taken to 16 hours. 

The TextFooler part consists of two steps-- the fine tuning part and attack part-- for generating adversaries. For fine tuning models we use "TeslaV100-SXM2-16GB", and it takes 20-30 minutes to complete the process. For the attack part we use "Tesla V100-SXM2-32GB", which takes 2-3 hrs for completing 20k data samples. This estimate requires the cosine similarity matrix for word embeddings to be calculated before hand which takes around 1-2 hrs, but this step has to be done only if the word embeddings are modified. This is a rare task so we have kept this separated.

\subsection{Hyper Parameters}\label{supp3}

To look at the estimations of DQI and its variations, we have kept basic hyper-parameters fixed in the experiments. We keep the learning rate to 1e-5, the number of epochs during the experiments are varied from 2-3, the per gpu train batch and eval batch sizes vary from 8-64 samples (the results shown are with respect to a batch size of 8), adam epsilon is set to 1e-8, weight decay is set to 0, maximum gradient normalisation is set to 1, and maximum sequence length is set to 128.  For TextFooler the the semantic similarity is fixed to 0.5 uniformly for all the experiments shown in this paper.

Additionally, the variations and range in the DQI parameters are dataset specific, i.e., hyperparameters depend on the application task. \cite{Mishra2020DQIAG} design DQI as a generic metric to evaluate diverse benchmarks. However, the definitions of what constitutes high and low quality will vary depending on the application. For example, BiomedicaNLP might have lower tolerance levels for spurious bias than General NLP. Another case is in water quality-- cited as an inspiration for DQI by \cite{Mishra2020DQIAG}--  where the quality of water needed for irrigation is different than that of drinking or medicine. We can therefore say that the hyper-parameters in the form of boundaries separating high and low quality data (i.e., inductive and spurious bias) are dependent on applications.

\subsection{Related Work}\label{supp4}
\subsubsection{Sample Quality and Artifacts}

Data Shapley \cite{ghorbani2019data} has been proposed as a metric to quantify the value of each training datum to the predictor performance. However, the metric might not signify bias content, as the value of training datum is quantified based on predictor performance, and biases might favor the predictor. Moreover, this approach is model and task-dependent. VAIDA uses DQI (Data Quality Index), proposed by ~\cite{Mishra2020DQIAG}, to: (i) compute the overall data quality for a benchmark with $n$ data samples, and (ii) compute the impact of a new $(n+1)^{th}$ data sample. \cite{wang2020vibe} concurrently propose a tool for measuring and mitigating artifacts in image datasets.

Data Shapley \cite{ghorbani2019data} has been proposed as a metric to quantify the value of each training datum to the predictor performance. However, this approach is model and task dependent. More importantly, the metric might not signify bias content, as the value of training datum is quantified based on predictor performance, and biases might favor the predictor. VAIDA uses DQI (data quality index), proposed by ~\cite{Mishra2020DQIAG}, to: (i) compute the overall data quality for a benchmark with $n$ data samples, and (ii) compute the impact of a new $(n+1)^{th}$ data sample. The quality of individual features (aspects) of samples are evaluated based on decreasing presence of artifacts and increasing generalization capability. In a concurrent work \cite{wang2020vibe}, a tool for measuring and mitigating bias in image datasets has also been proposed. DQI estimates artifact presence by calculating seven component values corresponding to a set of language properties;, along with their interpretation in VAIDA.

\subsubsection{Crowdsourcing Pipelines}
\label{sub:pipelines}

\textbf{Adversarial Sample Creation: } Pipelines such as Quizbowl\cite{wallace2019trick} and Dynabench\cite{kiela2021dynabench}, highlight portions of text from input samples during crowdsourcing, based on how important they are for model prediction; this prompts users to alter their samples, and produce samples that can fool the model being used for evaluation. While these provide more focused feedback compared to adversarial pipelines like ANLI \cite{nie2019adversarial}, which do not provide explicit feedback on text features, adversarial sample creation is contingent on performance against a specific model (Quizbowl for instance is evaluated against IR and RNN models, and may therefore not see significant performance drops against more powerful models). Additionally, such sample creation might introduce new artifacts over time into the dataset and doesn't always correlate with high quality-- for instance, a new entity introduced to fool a model in an adversarial sample might be the result of insufficient inductive bias, though reducing the level of spurious bias.

A similar diagnostic approach is followed for unknown unknown identification-- i.e., instances for which a model makes a high conﬁdence prediction that is incorrect. \cite{attenberg2015beat} and \cite{vandenhof2019hybrid} propose techniques to identify UUs, in order to discover specific areas of failure in model generalization through crowdsourcing. The detection of these instances is however, model-dependent; VAIDA addresses the occurrence of such instances by comparing sample characteristics between different labels to identify (and resolve) potential artifacts and/or under-represented features in created data. 

\textbf{Promoting Sample Diversity: } Approaches focusing on improving sample diversity have been proposed, in order to promote model generalization. \cite{yaghoub2020dynamic} use a probablistic model to generate word recommendations for crowdworker paraphrasing. \cite{larson2019outlier} propose retaining only the top k\% of paraphrase samples that are the greatest distance away from the mean sentence embedding representation of all collected data. These `outlier' samples are then used to seed the next round of paraphrasing. \cite{larson2020iterative} iteratively constrain crowdworker writing by using a taboo list of words, that prevents the repetition of over-represented words, which are also a source of spurious bias. Additionally, \cite{stasaski2020more} assess the new sample's contribution to the diversity of the entire sub-corpus.

DQI encompasses the aspects of artifacts studied by the aforementioned works; it further quantifies the presence of many more inter and intra-sample artifacts, and provides a one stop solution to address artifact impact on multiple fronts. VAIDA leverages DQI to identify artifacts, and further focuses on educating crowdworkers on exactly `why' an artifact is undesirable, as well as the impact its presence will have on the overall corpus. This is in contrast to the implicit feedback provided by word recommendation and/or highlighting in prior works-- VAIDA facilitates the elimination of artifacts without the unintentional creation of new artifacts, something that has hitherto remained unaddressed.

\subsubsection{Task Selection and Controlled Dataset Creation}

 In this work, we demonstrate VAIDA for a natural language inference task (though it is task-independent), and mimic the SNLI dataset creation and validation processes. Elicited annotation has been found to lead to social bias in SNLI using probablistic mutual information (PMI) \cite{rudinger2017social}. Visual feedback is provided based on DQI (which takes PMI into account) to explicitly correct this bias, and discourage the creation of such samples. Also, human annotation of machine-generated sentences/sentences pulled from existing texts instead of elicitation has been suggested to reduce such bias \cite{zhang2017ordinal}. However, machine-generated text might look artificial, and work has shown that text generation has its own set of quality issues \cite{mathur2020tangled}. While we use AutoFix and TextFooler as modules to automatically transform samples, they are designed to be used in parallel with human sample creation. Their results can also be further modified by humans prior to submission. We see less reliance on these tools over the course of our user study, as discussed in Subsection 6.2. Additionally, previous work \cite{roit2019crowdsourcing} in controlled dataset creation trains crowdworkers, and selects a subset of the best-performing crowdworkers for actual corpus creation. Each crowdworker's work is reviewed by another crowdworker, who acts as an analyst (as per our framework) of their samples. However, in real-world dataset creation, such training and selection phases might not be possible. Additionally, the absence of a metric-in-the-loop basis for feedback provided during training can potentially bias (through trainers) the created samples.
 
 \subsection{DQI Components}\label{supp5}

DQI shows the (i) the overall data quality and (ii) the impact of new data created on the overall quality. In this paper, higher quality implies lower artifact presence and higher generalization capability. DQI clubs artifacts into seven broad aspects of text, which cover the space of various possible interactions between samples in an NLP dataset. Please refer to \cite{Mishra2020DQIAG} and \cite{mishra2022survey} for a full explanation of parameters. 

\begin{figure*}[!t]
\includegraphics[width=\linewidth,height=8.55cm]{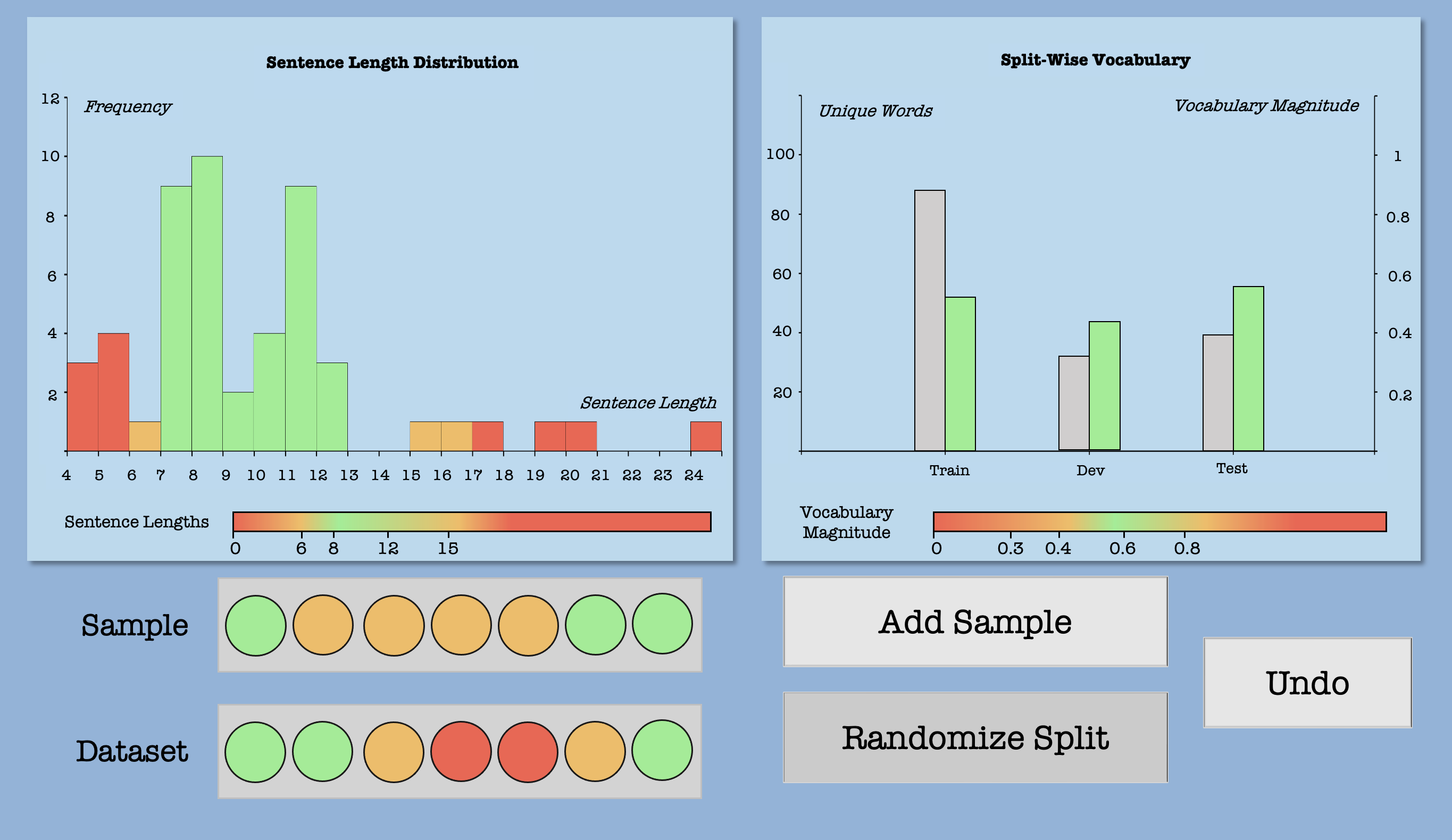}
  \caption{$DQI_{c1}$ Visualization Prior to New Sample Addition}
  \label{fig:Vis1before}
\includegraphics[width=\linewidth,height=8.55cm]{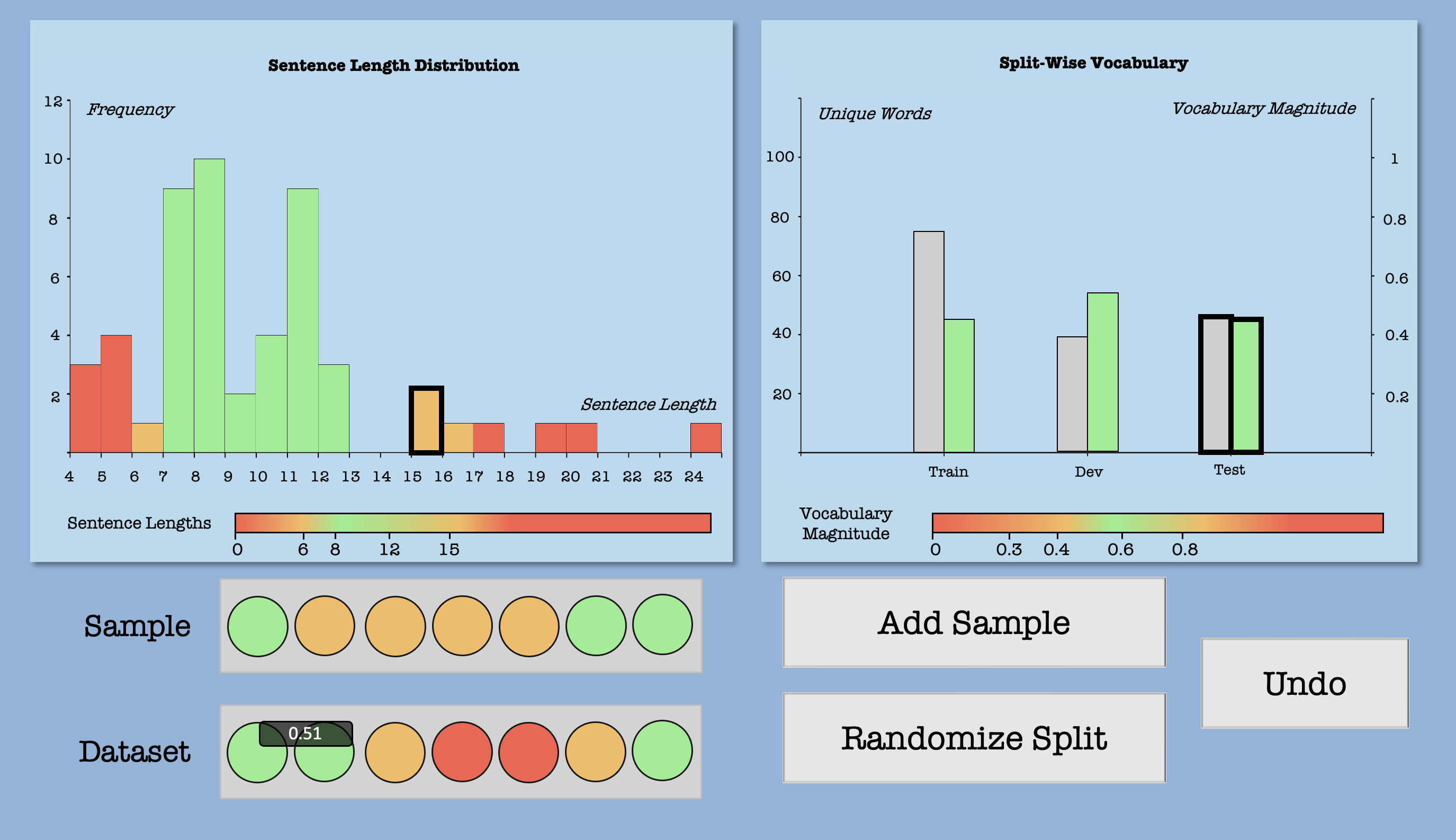}
  \caption{$DQI_{c1}$ Visualization On New Sample Addition}
  \label{fig:Vis1after}
\end{figure*}

\subsection{Interface Design Intuitions}\label{supp7}

\paragraph{Careful Selection of Visualizations}
Prior to the design of test cases and a user interface, data visualizations highlighting the effects of sample addition are built. Considering the complexity of the formulas for the components of empirical DQI, we carefully select visualizations to help illustrate and analyze the effect to which individual text properties are affected. 

\subsubsection{Vocabulary}
\paragraph{Which Characteristics of Data are Visualized?}
The contribution of samples to the size of the vocabulary is tracked using a dual axis bar chart. This displays the vocabulary size, along with the vocabulary magnitude, across the train, dev, and test splits for the dataset. By randomizing data splits, the distribution of vocabulary across dataset samples can clearly be identified. We use a dual axis chart as juxtaposition of the vocabulary magnitude against raw counts of words better reflects the evenness of the vocabulary distribution; it is not useful to have only a few samples contributing new words as other samples automatically become easy for the model to solve.

To further clarify the contribution of individual samples to vocabulary, the distribution of sentence lengths is plotted as a histogram. Each sample contributes two sentences, i.e., the premise and hypothesis statements. Figure \ref{fig:Vis1before} illustrates this. The histogram provides analysts with a frame of reference to identify gaps or outliers in the distribution, essential for determining which sentences are undesirable for the corpus due to extremely high (low inductive bias) or extremely low (high artifact-- spurious bias) vocabulary contribution.

\subsubsection{Inter-sample N-gram Frequency and Relation}
\begin{figure*}[!t]
\includegraphics[width=\linewidth,height=8.55cm]{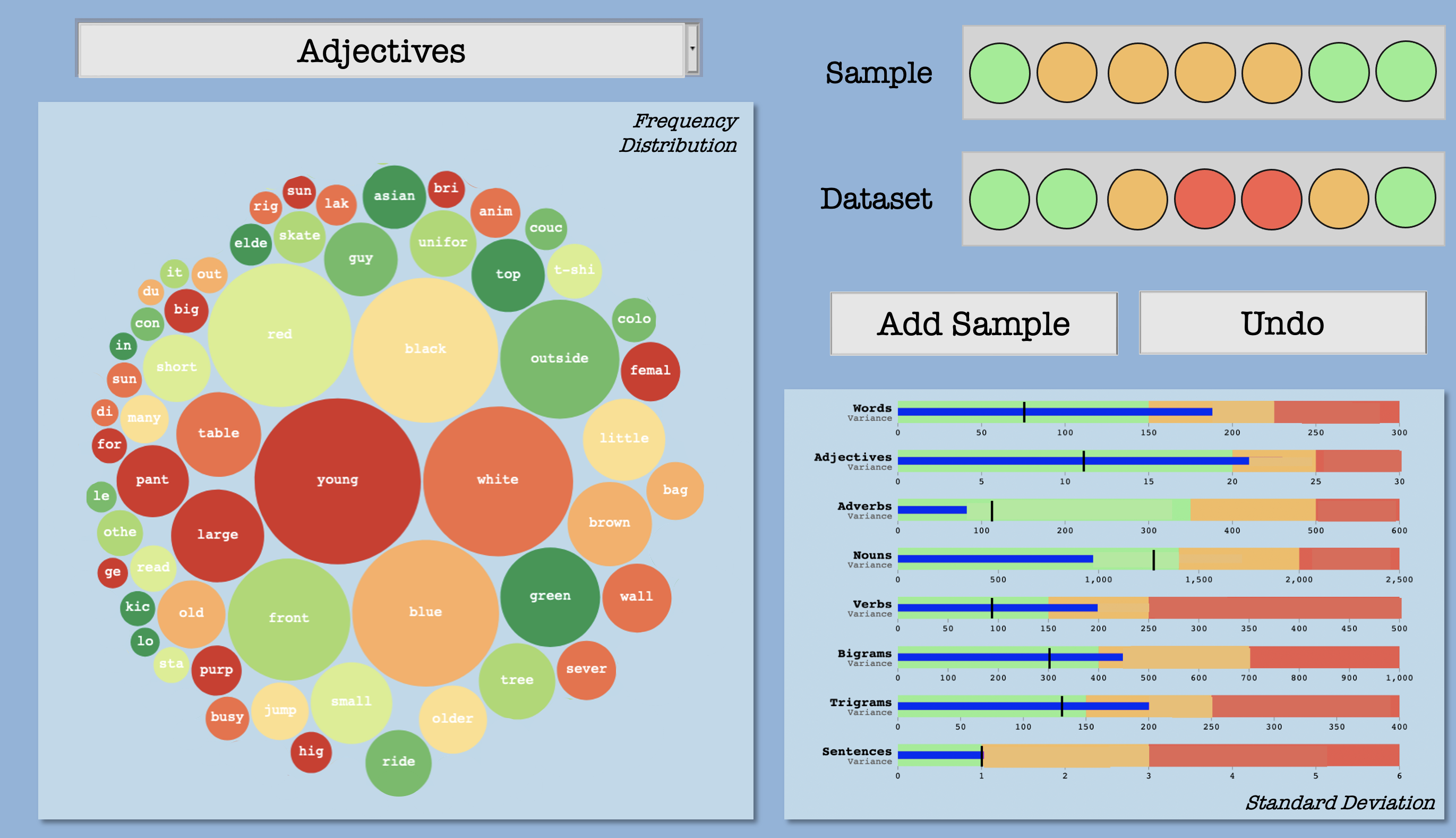}
  \caption{$DQI_{c2}$ Visualization Prior to New Sample Addition}
  \label{fig:Vis2before}
\includegraphics[width=\linewidth,height=8.55cm]{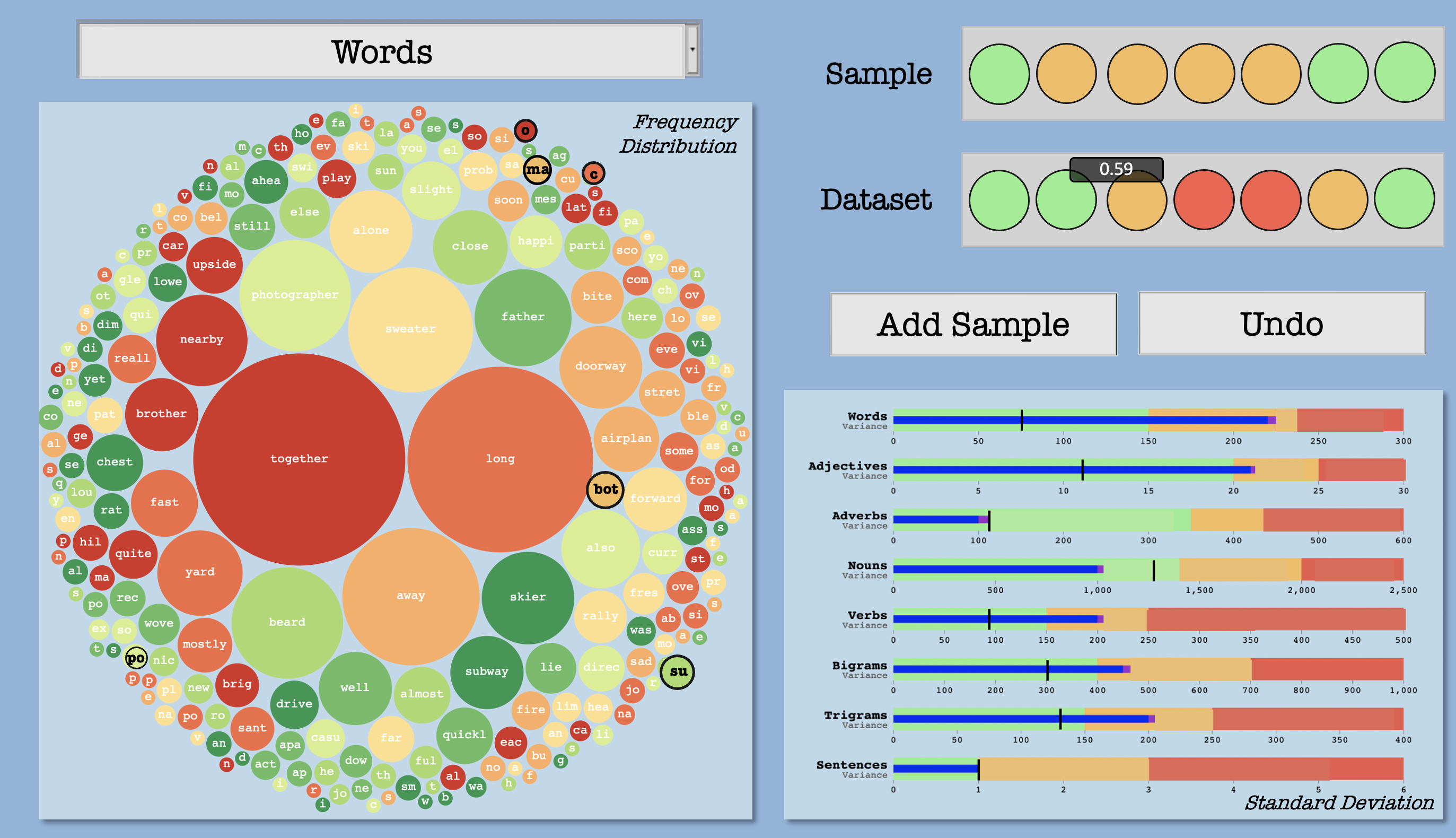}
  \caption{$DQI_{c2}$ Visualization On New Sample Addition}
  \label{fig:Vis2after}
\end{figure*}

\paragraph{Which Characteristics of Data are Visualized?}
There are different granularities of samples that are used to calculate the values of this component, namely: words, POS tags, sentences, bigrams, and trigrams. The granularities' respective frequency distributions and standard deviations are utilized for this calculation.
\paragraph{Bubble Chart for visualizing the frequency distribution:}
A bubble chart is used to visualize the frequency distribution of the respective granularity. This design choice is made in order to clearly view the contribution made by a new sample when added to the existing dataset in terms of different granularities. The bubbles are colored according to the bounds set for frequencies by the hyperparameters, and sized based on the frequency of the elements they represent. Additionally, some insight into variance can be obtained from this chart, by observing the variation in bubble size. 
\paragraph{Bullet Chart for impact of new sample:}
The impact of sample addition on standard deviation can be viewed using the bullet chart. The bullet chart is useful to visually track performance against a target (in this case ideal standard deviation), displaying results in a single column; it looks like a thermometer and is therefore easy to follow. The red-yellow-green color bands for each granularity represent the standard deviation bounds of that granularity. The vertical black line represents the ideal value of the standard deviation of that granularity. The two horizontal bars represent the value of standard deviation before and after the new sample's addition. Figure \ref{fig:Vis2before} illustrates the visualization.

\begin{figure*}[!t]
\includegraphics[width=\linewidth,height=8.55cm]{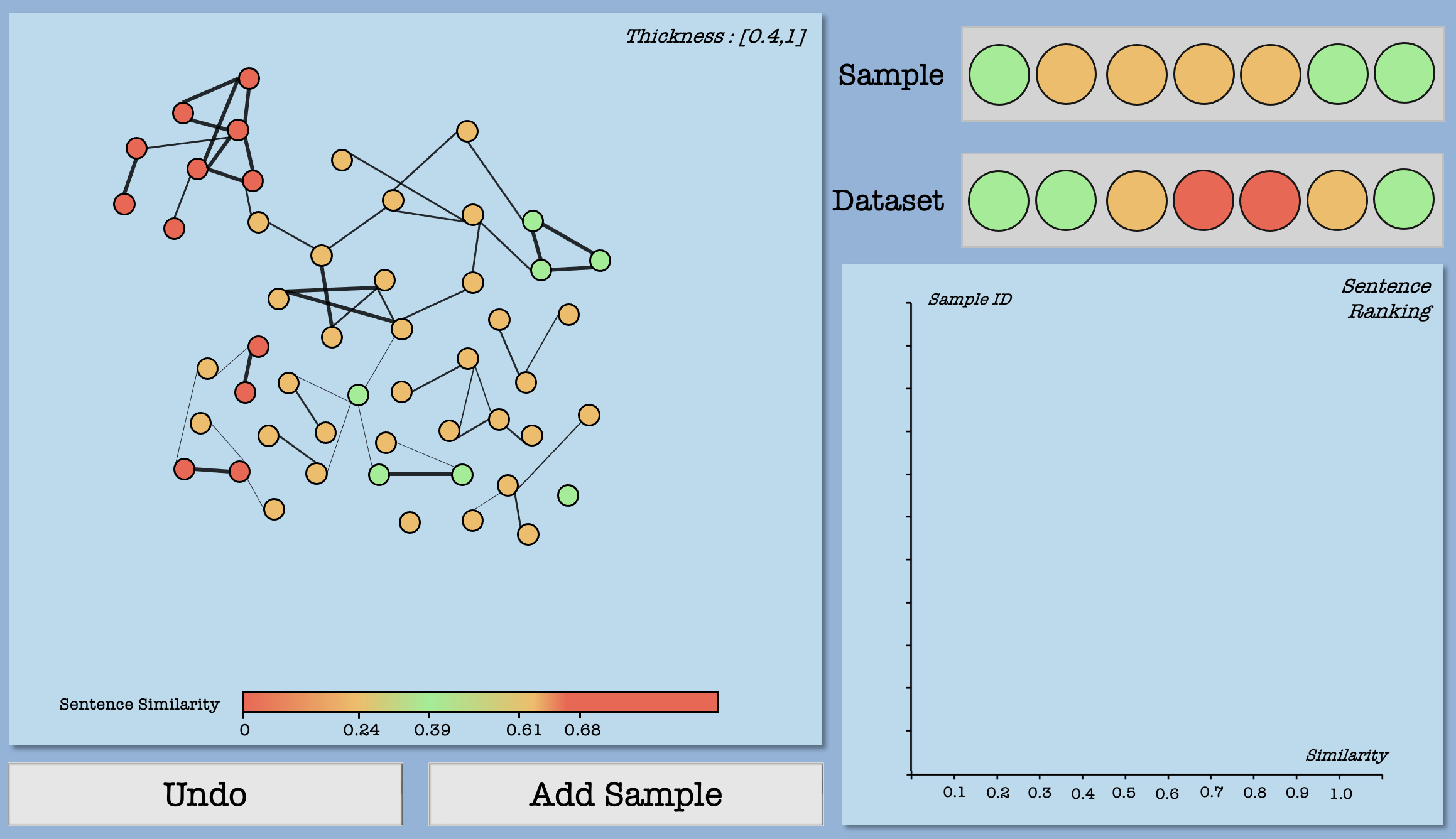}
  \caption{$DQI_{c3}$ Visualization Prior to New Sample Addition}
  \label{fig:Vis3before}
\includegraphics[width=\linewidth,height=8.55cm]{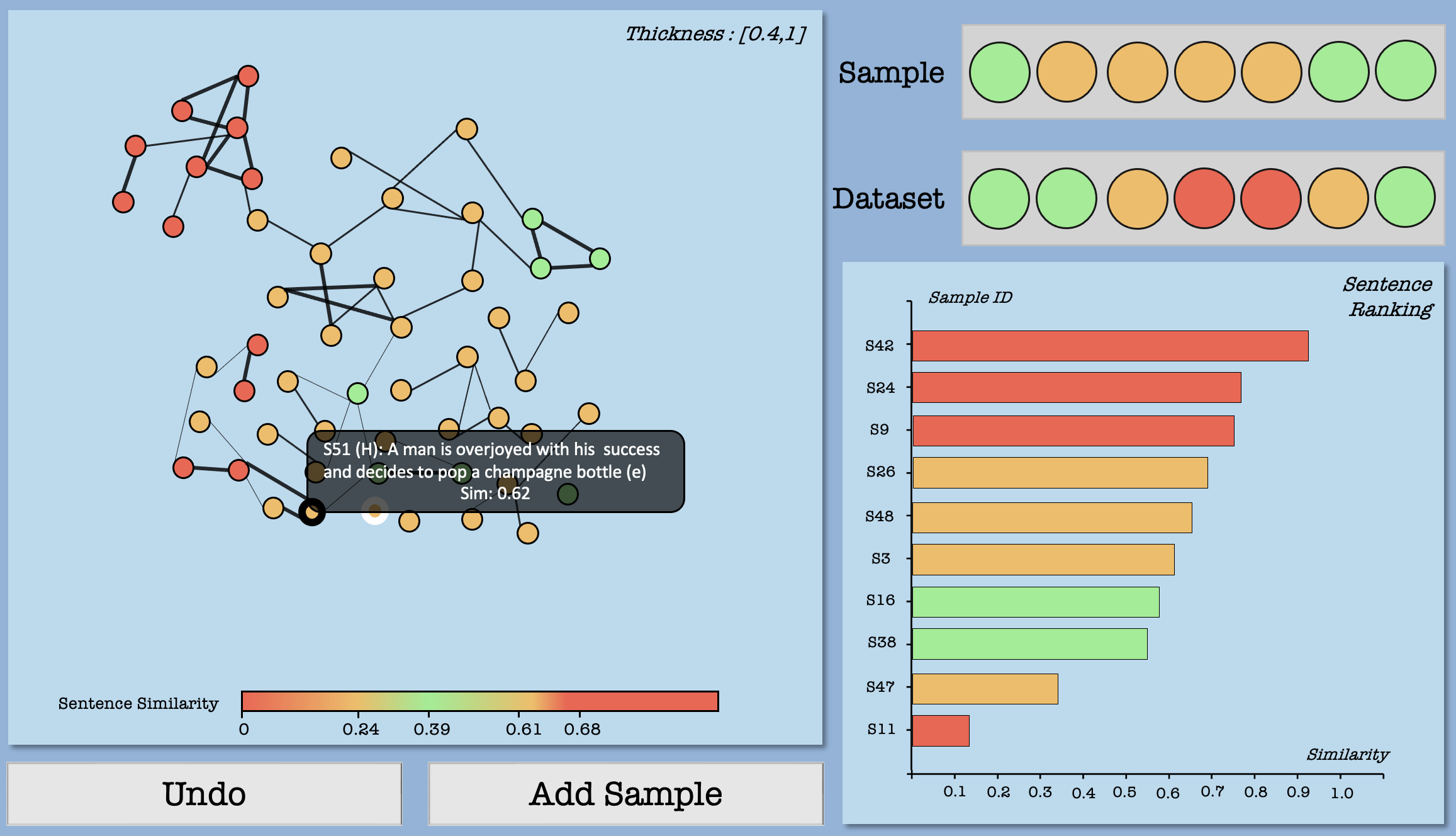}
  \caption{$DQI_{c3}$ Visualization On New Sample Addition}
  \label{fig:Vis3after}
\end{figure*}

\subsubsection{Inter-sample STS}
\paragraph{Which Characteristics of Data are Visualized?}
The main units used in this DQI component are the similarity values between sentences across the dataset. This refers to either premise or hypothesis statements, relative to all other premise/hypothesis statements.  In order to understand the similarity relations of sentences, a force layout and horizontal bar chart are used. This is illustrated in Figure \ref{fig:Vis3before}.
\paragraph{Force Layout for Similar Sentence Pairs}
In the force layout, those sentence pairs with a similarity value that meets the minimum threshold are connected. Each node represents a sentence. The thickness of the connecting line depends on how close the similarity value is to the threshold. Similarity values are used to create this network, as the aim of this component is to drill down into whether a sample has sufficient inductive bias (i.e., is closely linked to a sufficient extent to exiting samples), and also if a sample is too similar (spurious bias) to existing samples.
\paragraph{Horizontal Bar Chart for Most Similar Sentences}
In the horizontal bar chart, the sentences that are most similar to the given sentence are ordered in terms of their similarity value. The bar colors are centered around the threshold. This helps identify the most important subset to juxtapose the given sample against;the analyst can use this subset to for instance, decide if a moderate quality sample requires a small or big change in order to reacch acceptable quality..

\subsubsection{Intra-sample Word Similarity}
\paragraph{Which Characteristics of Data are Visualized?}
In this section, A sample's word similarity is viewed in terms
of premise-only, hypothesis-only, and both. The relationship between non-adjacent words in the sample's sentences is analyzed specifically.

\begin{figure*}[!t]
\includegraphics[width=\linewidth,height=8.55cm]{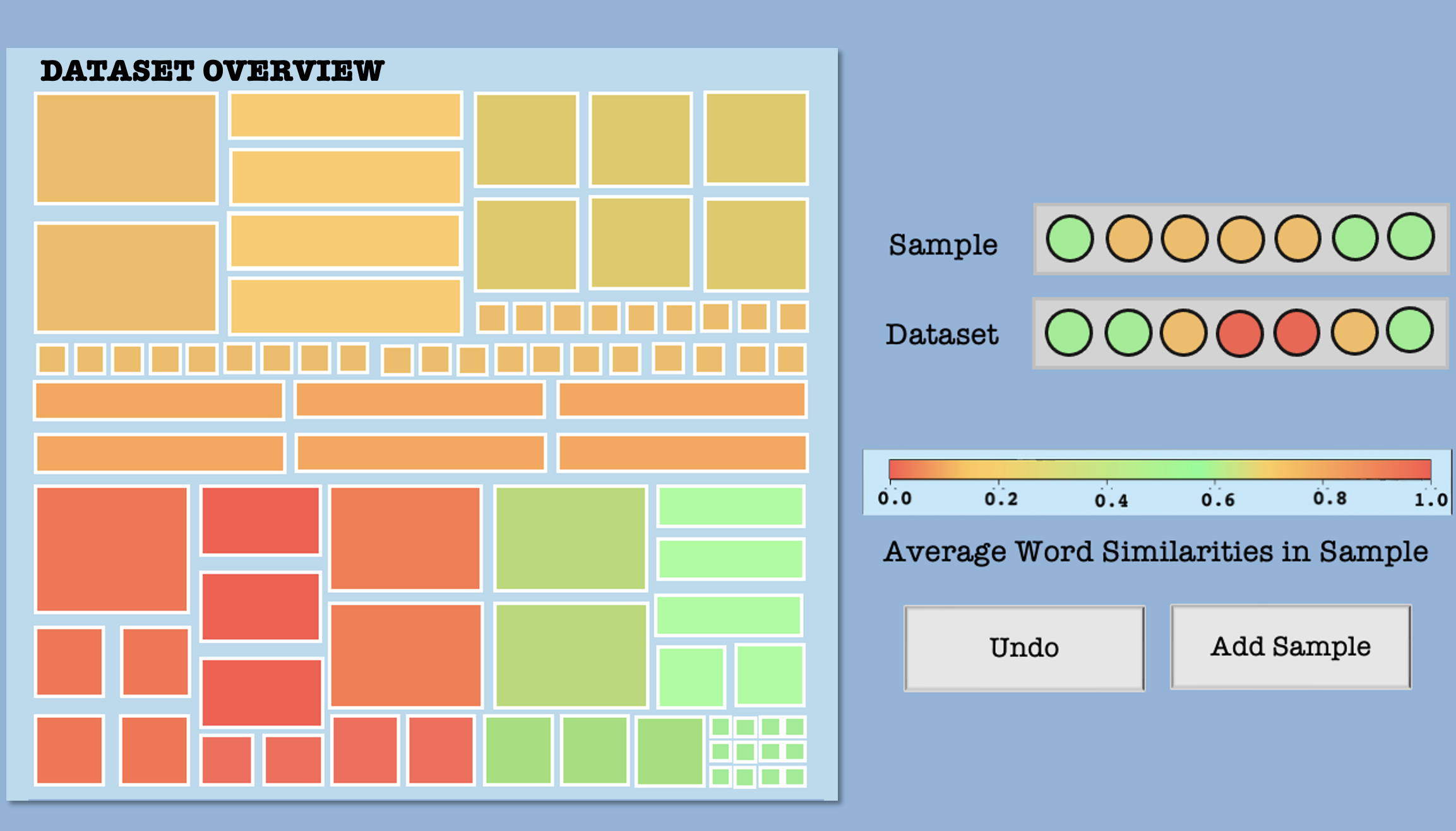}
  \caption{$DQI_{c4}$ Visualization Prior to New Sample Addition}
  \label{fig:Vis4before}
\includegraphics[width=\linewidth,height=8.55cm]{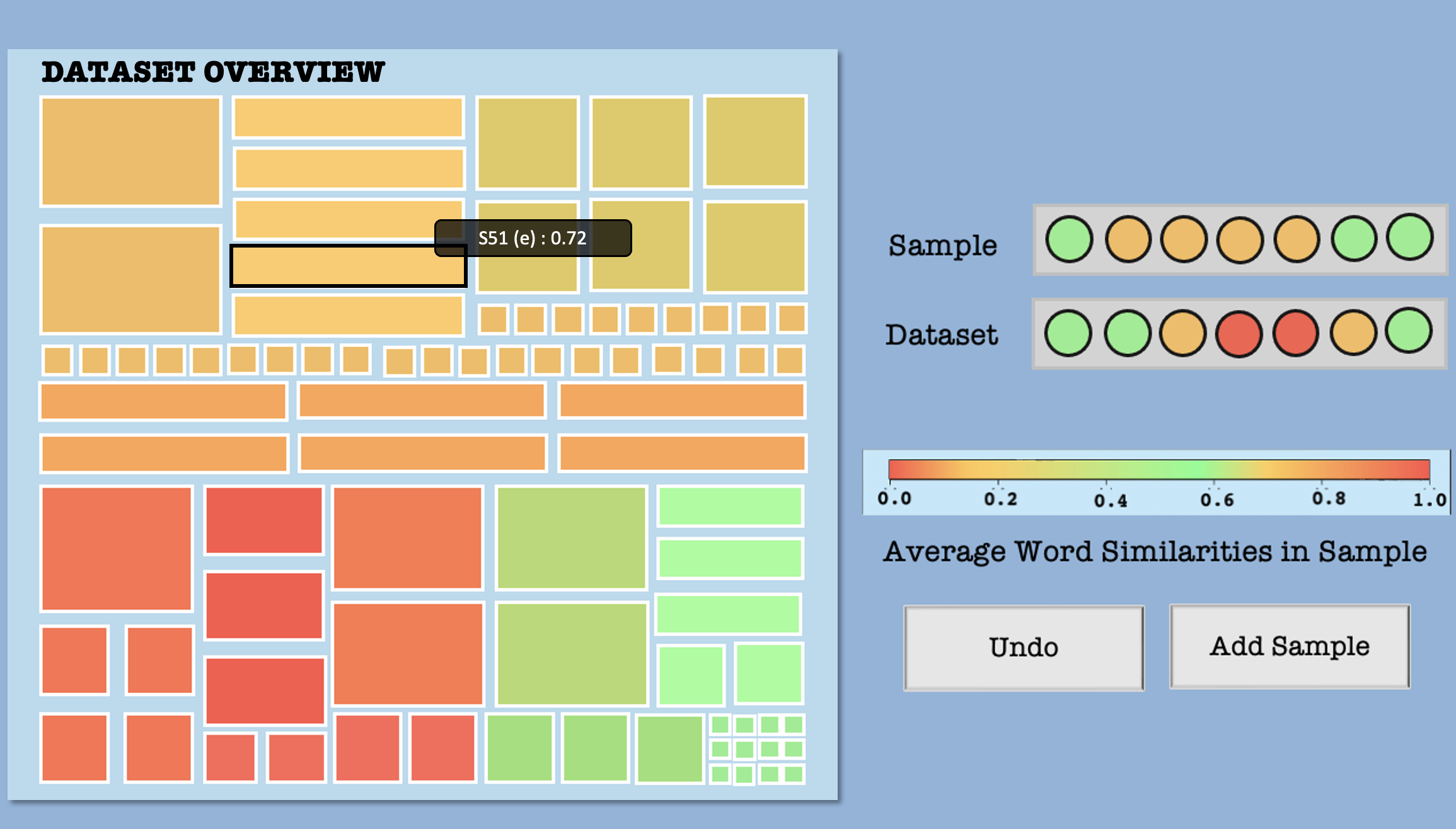}
  \caption{$DQI_{c4}$ Visualization On New Sample Addition: Dataset View}
  \label{fig:Vis4after}
\end{figure*}

\paragraph{Overview Chart for Average Word Similarities and Heatmap for Single Sample}
The overview chart that is used is a tree map, which uses the average value of all word similarities per sample- i.e., concatenated premise and hypothesis- to color and group its components. This is illustrated in Figure \ref{fig:Vis4before}. Treemaps capture relative sizes of data categories, allowing for quick perception of the items that are large contributors to each category. This makes them ideal to analyze the inter-relationships between different word pairs across sample, in a concise manner.

The treemap also makes it easy to drill down into the specifics of a particular sample even further. This detailed view is provided in the form of a heatmap. All the words in a single sample, are plotted against each other, as shown in Figure \ref{fig:Vis4component}. The heatmap provides a mechanism for word-level drill down of sample similarity. Like with the previous component, this helps provide the analyst with a frame of reference as to whether a moderate quality sample can be sent to TextFooler or not.

\subsubsection{Intra-sample STS}
\begin{figure*}[!t]
\includegraphics[width=\linewidth,height=8.55cm]{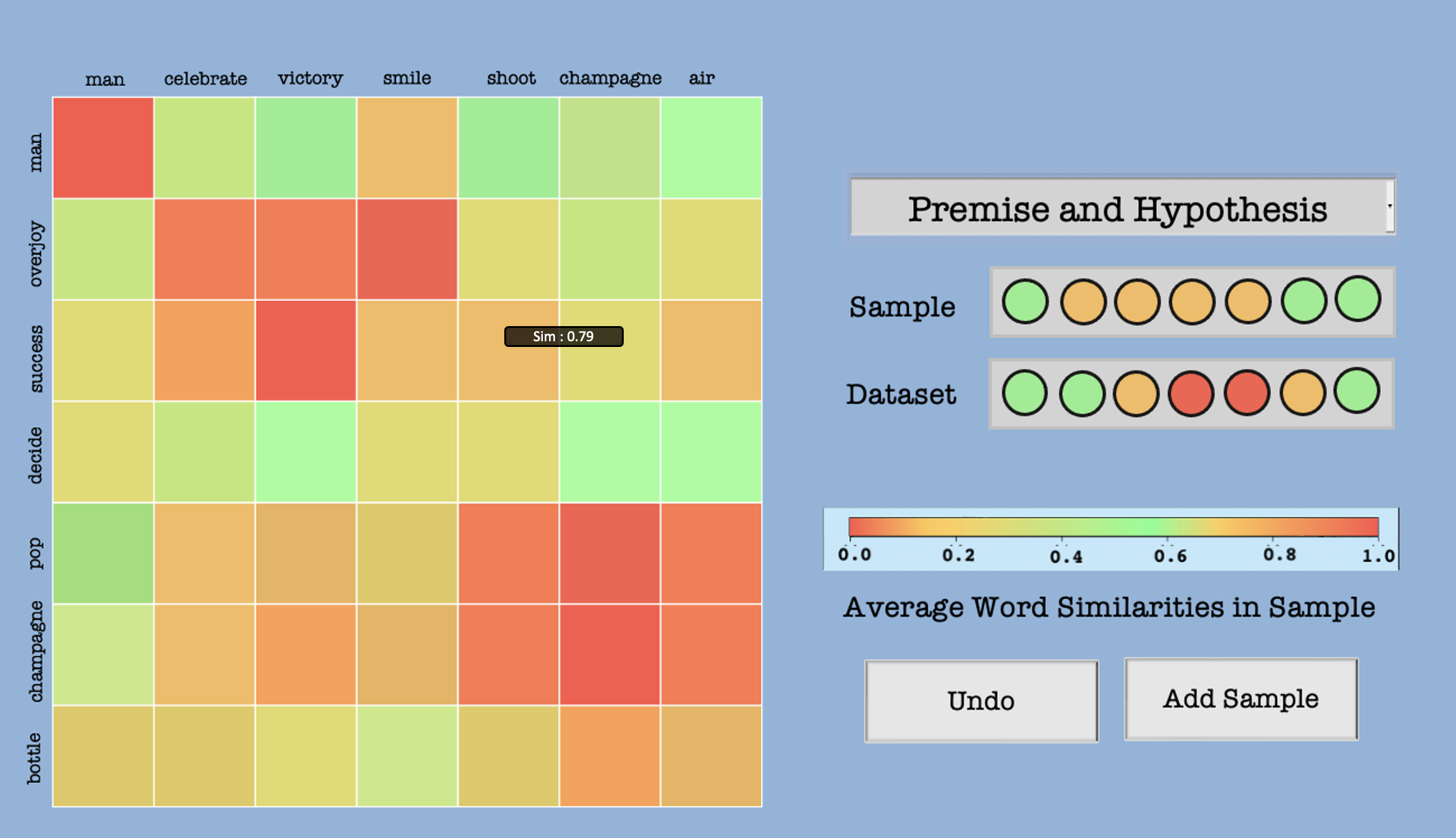}
  \caption{$DQI_{c4}$ Visualization On New Sample Addition: Sample View}
  \label{fig:Vis4component}
\includegraphics[width=\linewidth,height=8.55cm]{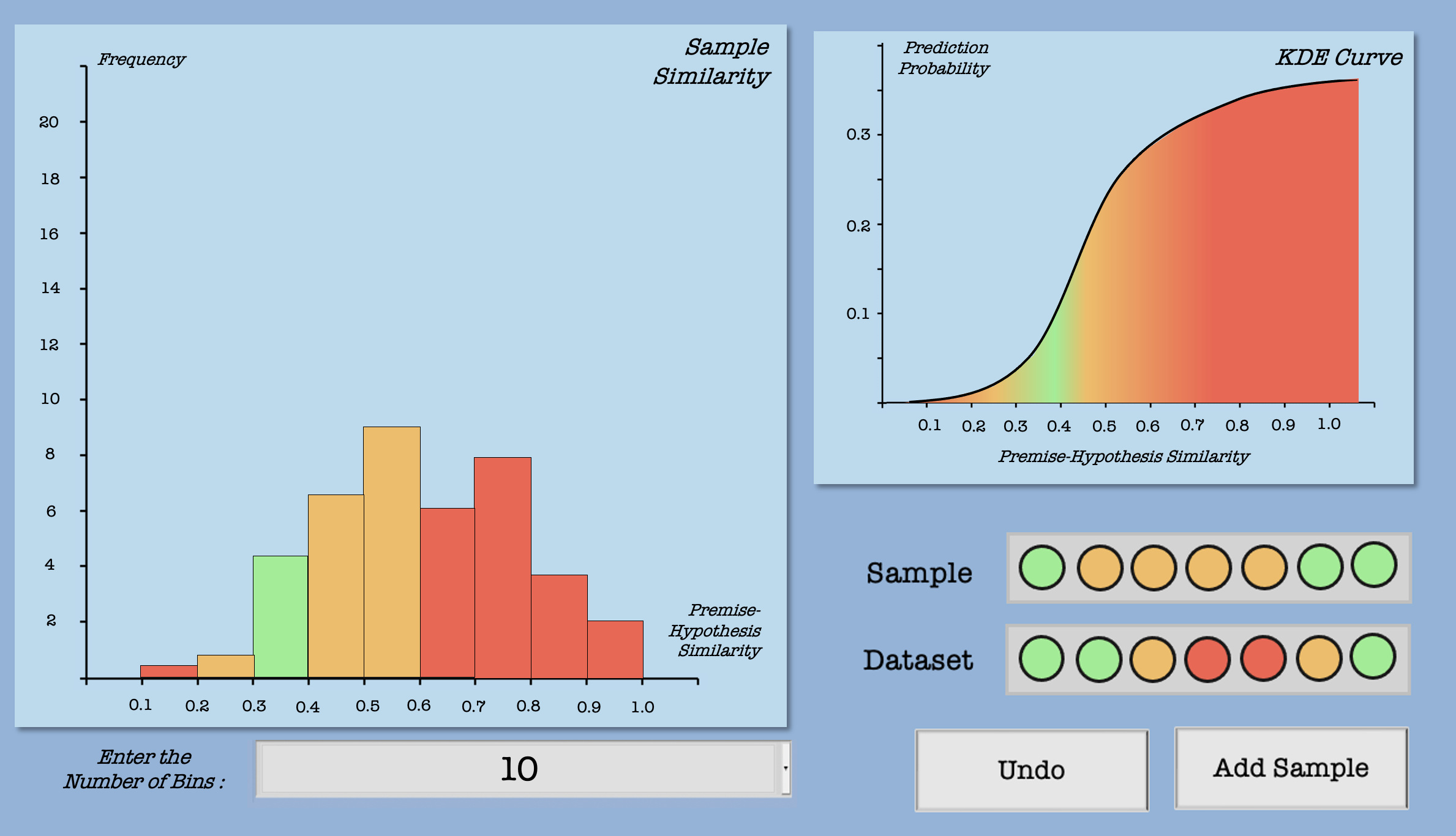}
  \caption{$DQI_{c5}$ Visualization Prior to New Sample Addition}
  \label{fig:Vis5before}
\end{figure*}

\paragraph{Which Characteristics of Data are Visualized?}
Premise-Hypothesis similarity is analyzed on the basis of length variation, meeting a minimum threshold, and similarity distribution across the dataset. The first is addressed already in the vocabulary property by viewing the sentence length distribution. The other two are visualized using a histogram and kernel density estimation curve, as shown in Figure \ref{fig:Vis5before}.
\paragraph{Histogram and Kernel Density Curve for Sample Distribution}
The histogram represents the distribution of the samples, and is colored by centering around the threshold as the ideal value. The number of bins can be changed, and therefore multi-level analysis can be conducted. This helps identify the proportion of samples that are of non-optimal range.

The kernel density curve is used to check for the overall skew of the distribution. KDE helps visualize the distribution sans user defied bins; this is colored to reflect the bi-linear color scale, and indicates the probability with which a sample will be correctly solved by a model given its similarity. This helps contextualize the level of artifacts relating to word overlap and sentence similarity in the sample.

\begin{figure*}[!t]
\includegraphics[width=\linewidth,height=8.55cm]{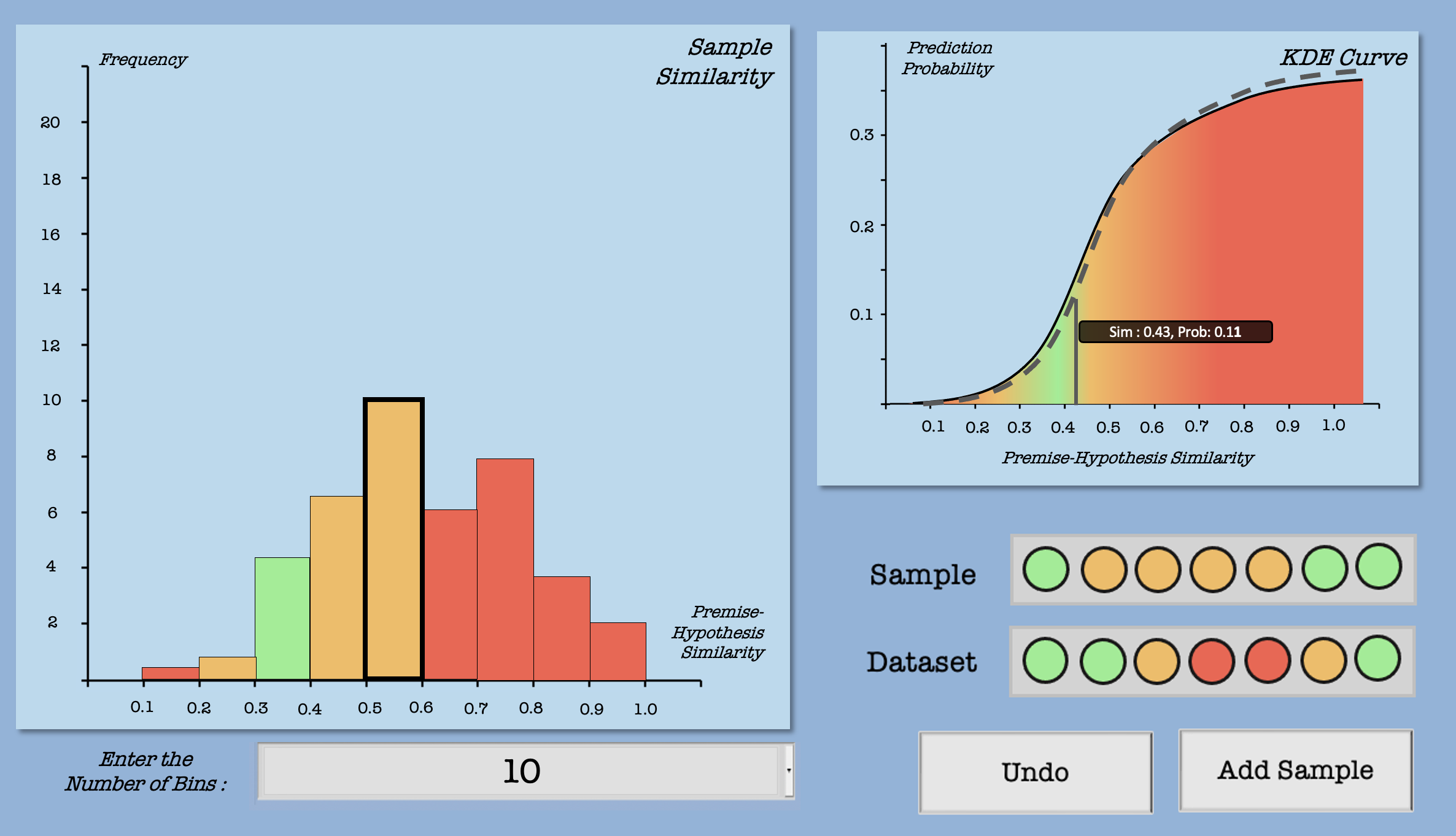}
  \caption{$DQI_{c5}$ Visualization On New Sample Addition}
  \label{fig:Vis5after}
\includegraphics[width=\linewidth,height=8.55cm]{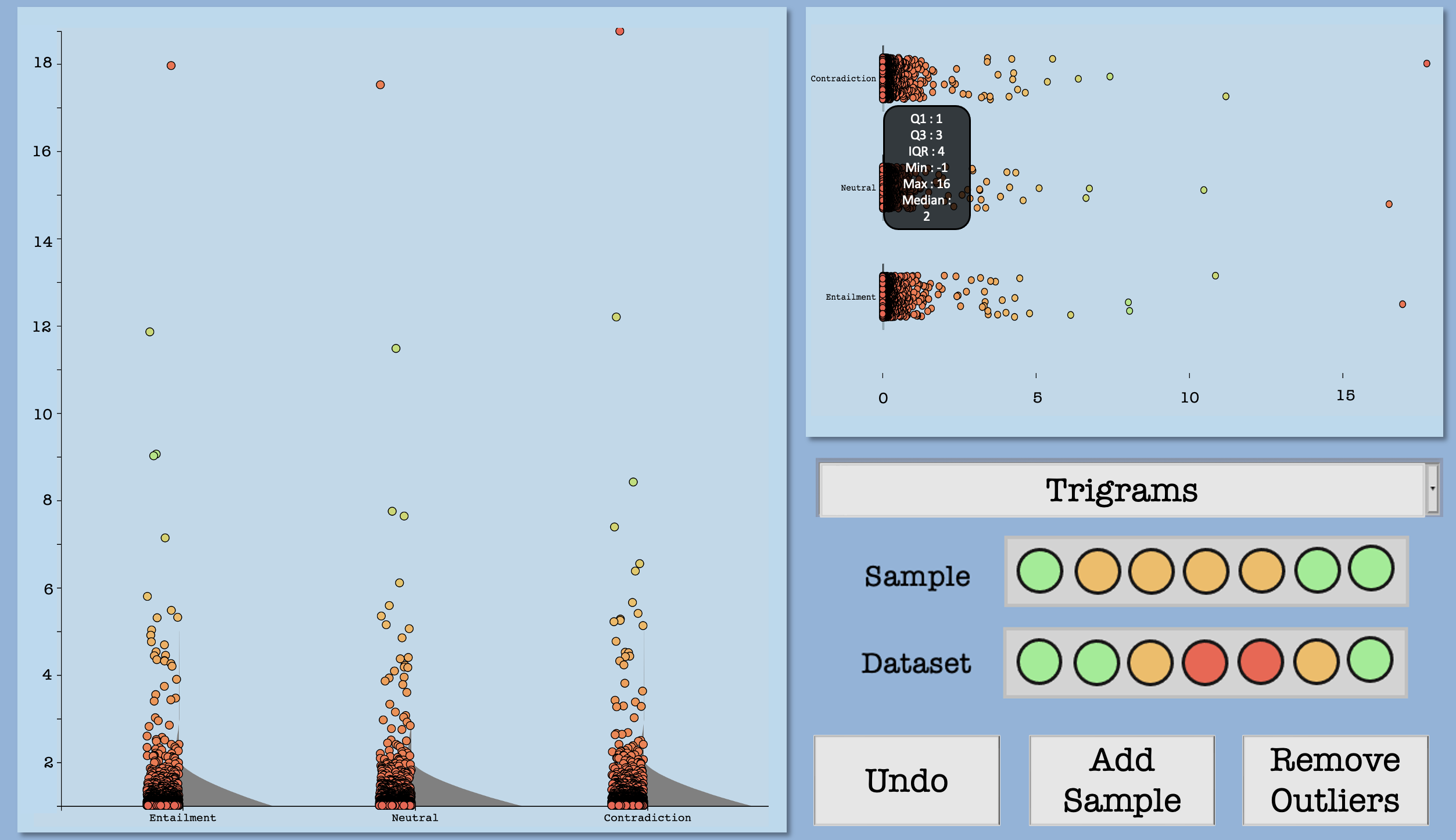}
  \caption{$DQI_{c6}$ Visualization Prior to New Sample Addition}
  \label{fig:Vis6before1}
\end{figure*}

\subsubsection{N-Gram Frequency per Label}
\paragraph{Which Characteristics of Data are Visualized?}
This component drills down on the second component, to view the patterns seen in granularities per label. There are two small multiples charts, divided based on label, used in this view- a violin plot and a box plot.
\paragraph{Violin plot and Kernel Density Curve for Skew of Distribution:}
The violin plots are structured to display both jittered points, according to their frequency distribution, as well as a kernel density curve to judge the skew of the distribution. The points each represent an element of the granularity. 
\paragraph{Box Plots for More Information}
The box plots are used to garner more information about the distribution, in terms of its min, max, median, mean, and inter quartile range. These help further characterize the distribution, as well as provide a quantitative definition of the skew seen using density curves. Jittered points representing elements are present in this plot as well.

\begin{figure*}[!t]
\includegraphics[width=\linewidth,height=8.55cm]{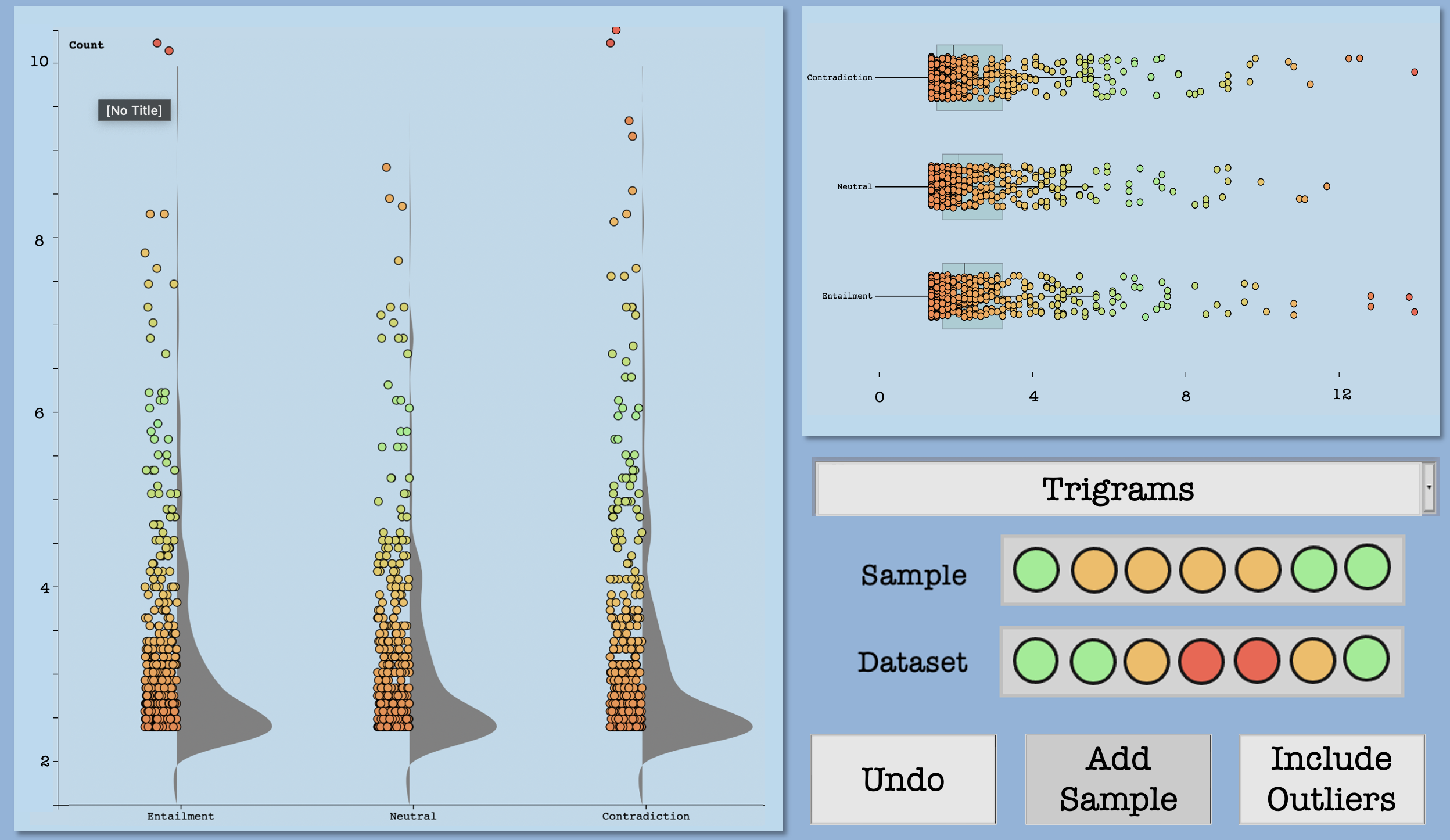}
  \caption{$DQI_{c6}$ Visualization after removing outliers Prior to New Sample Addition}
  \label{fig:Vis6before2}
\end{figure*}

\begin{figure*}[!t]
\includegraphics[width=\linewidth,height=8.55cm]{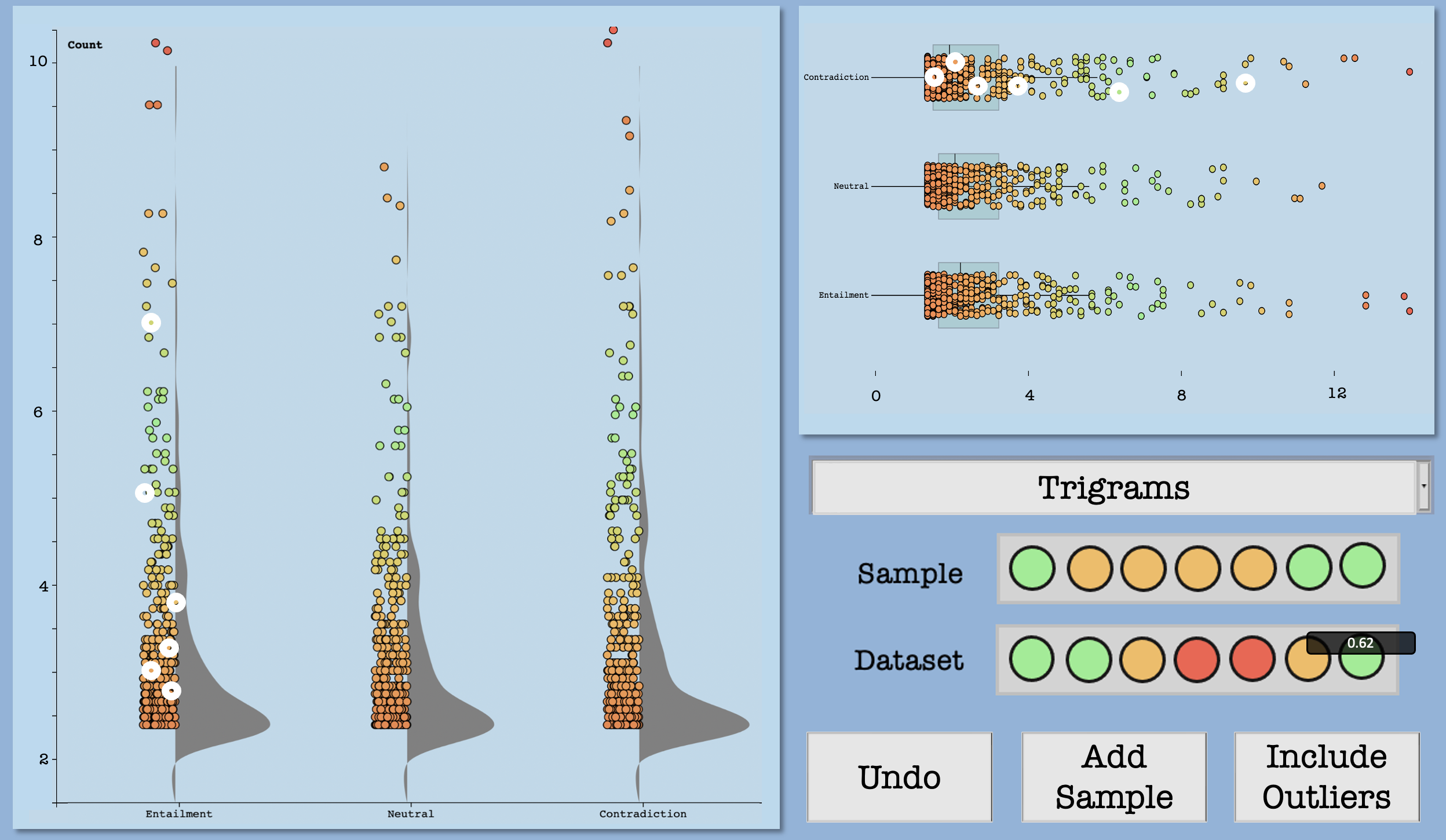}
  \caption{$DQI_{c6}$ Visualization with mouseover On New Sample Addition}
  \label{fig:Vis6after2}
\includegraphics[width=\linewidth,height=8.55cm]{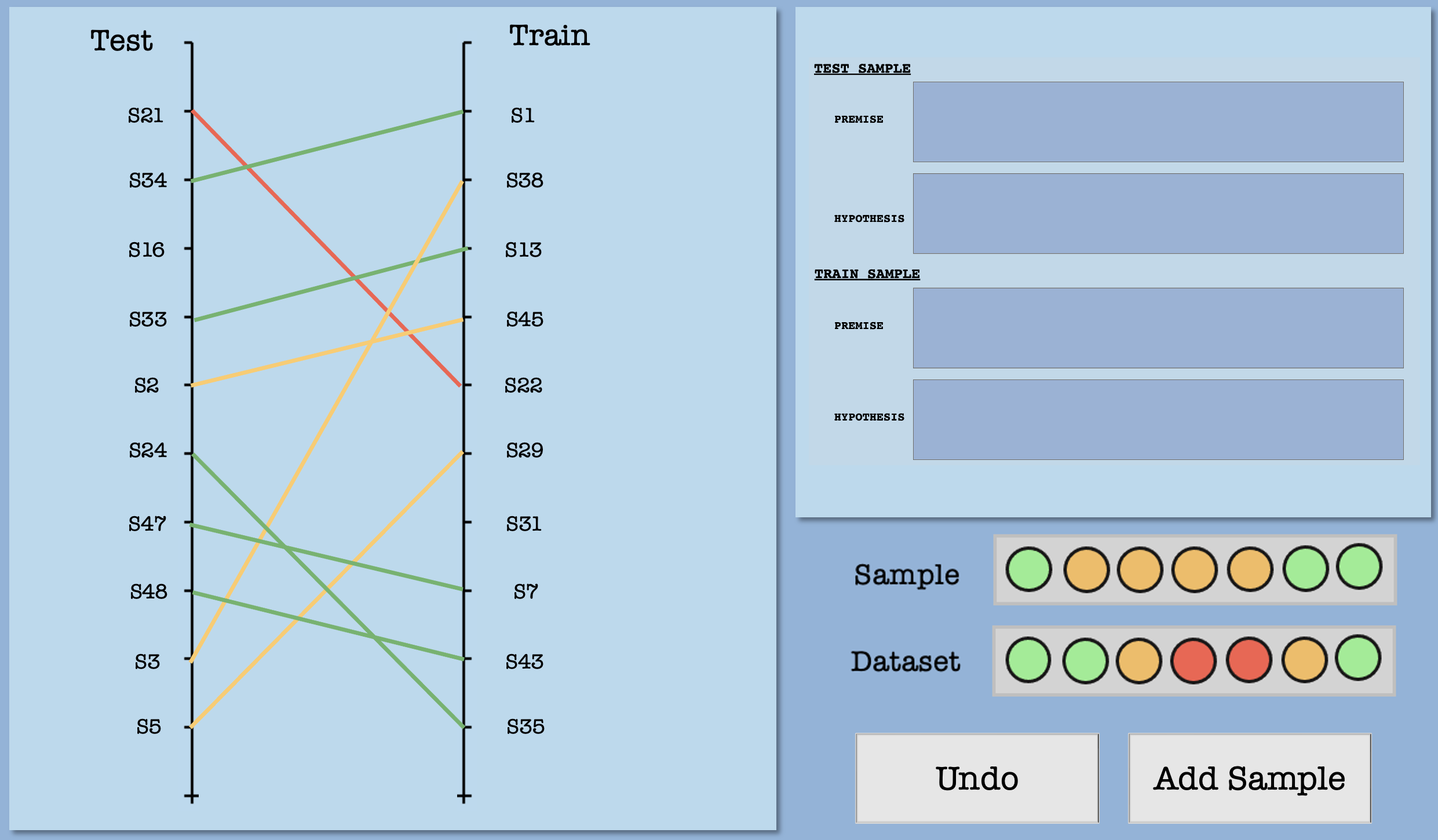}
  \caption{$DQI_{c7}$ Visualization Prior to New Sample Addition}
  \label{fig:Vis7before}
\end{figure*}

\begin{figure*}[!t]
\includegraphics[width=\linewidth,height=8.55cm]{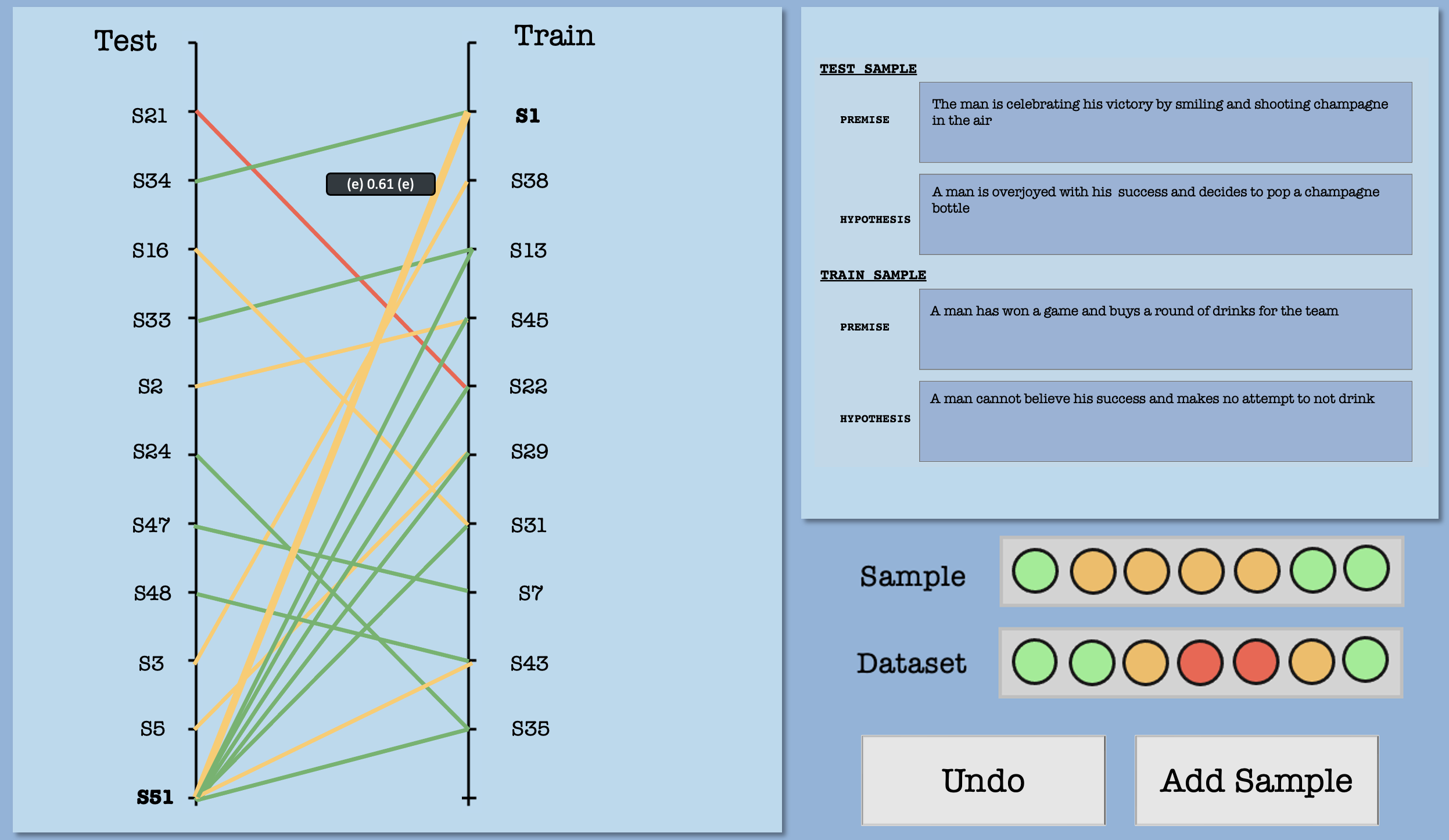}
  \caption{$DQI_{c7}$ Visualization On New Sample Addition}
  \label{fig:Vis7after}
\end{figure*}

\subsubsection{Inter-split STS}
\paragraph{Which Characteristics of Data are Visualized?}
Train-Test similarity must be kept minimal to prevent data leakage. This component's main feature is finding the train split sample that is most similar to a given test split sample.
\paragraph{Parallel Coordinate Graph for Train-Test Similarity:} A subset of test and train samples, all found to have close similarity within their respective splits, and significant similarity across the splits are plotted as a one step parallel coordinate graph, with test samples along one axis, and train samples along the other. This subset is seeded with those samples closest in similarity to the new sample to be introduced, based on the third component's visualization. The links connecting points on the two axes are drawn between the most similar matches across the split, as shown in Figure \ref{fig:Vis7before}. 

\subsection{AutoFix and TextFooler Examples}
\label{supp8}

See Tables \ref{tab:Autofixeg}, \ref{tab:tfeg}.

\begin{figure*}[t]
\includegraphics[width=\linewidth,height=8.55cm]{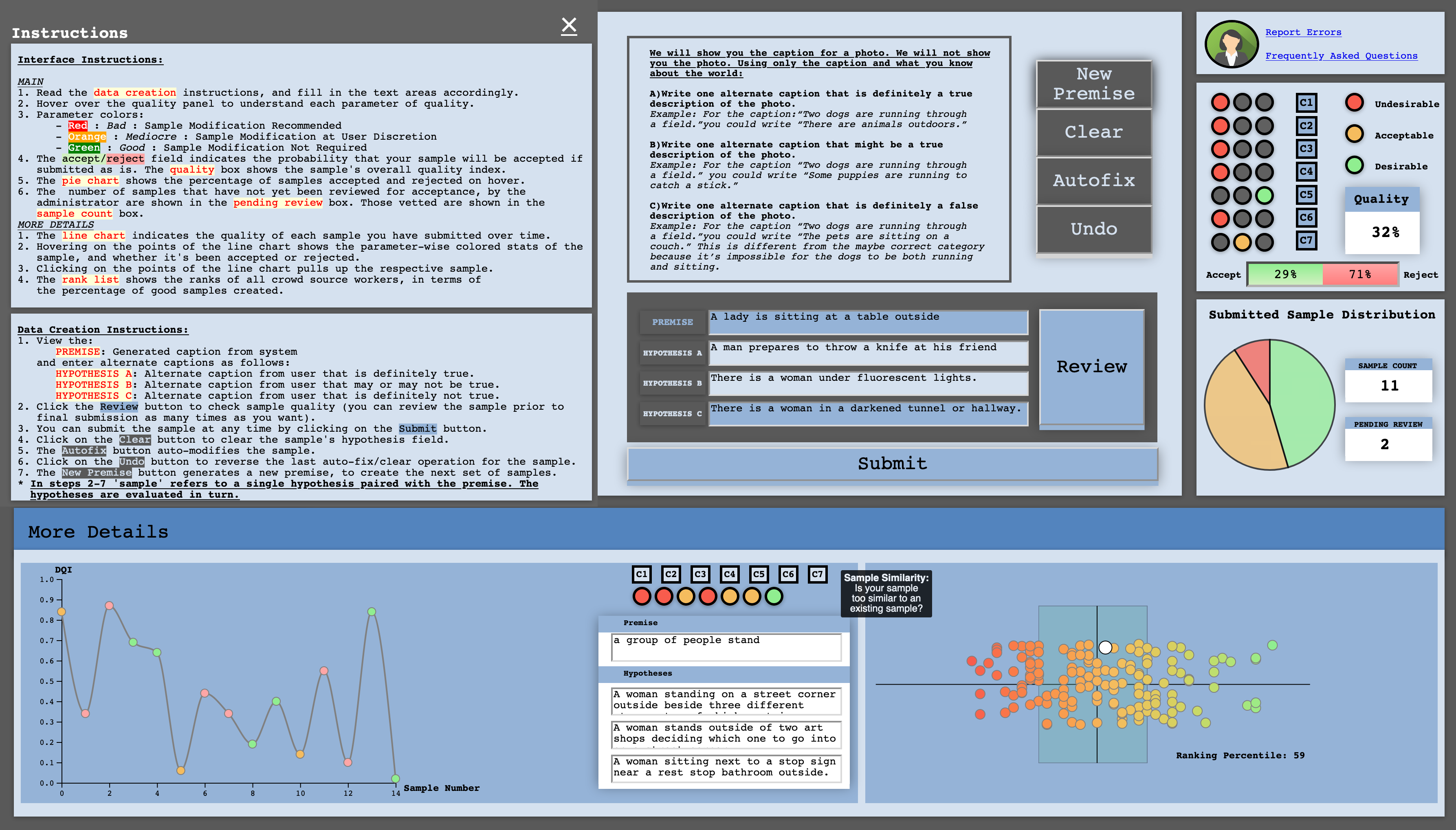}
  \caption{Crowdworker-View}
  \label{fig:cview}
\includegraphics[width=\linewidth,height=8.55cm]{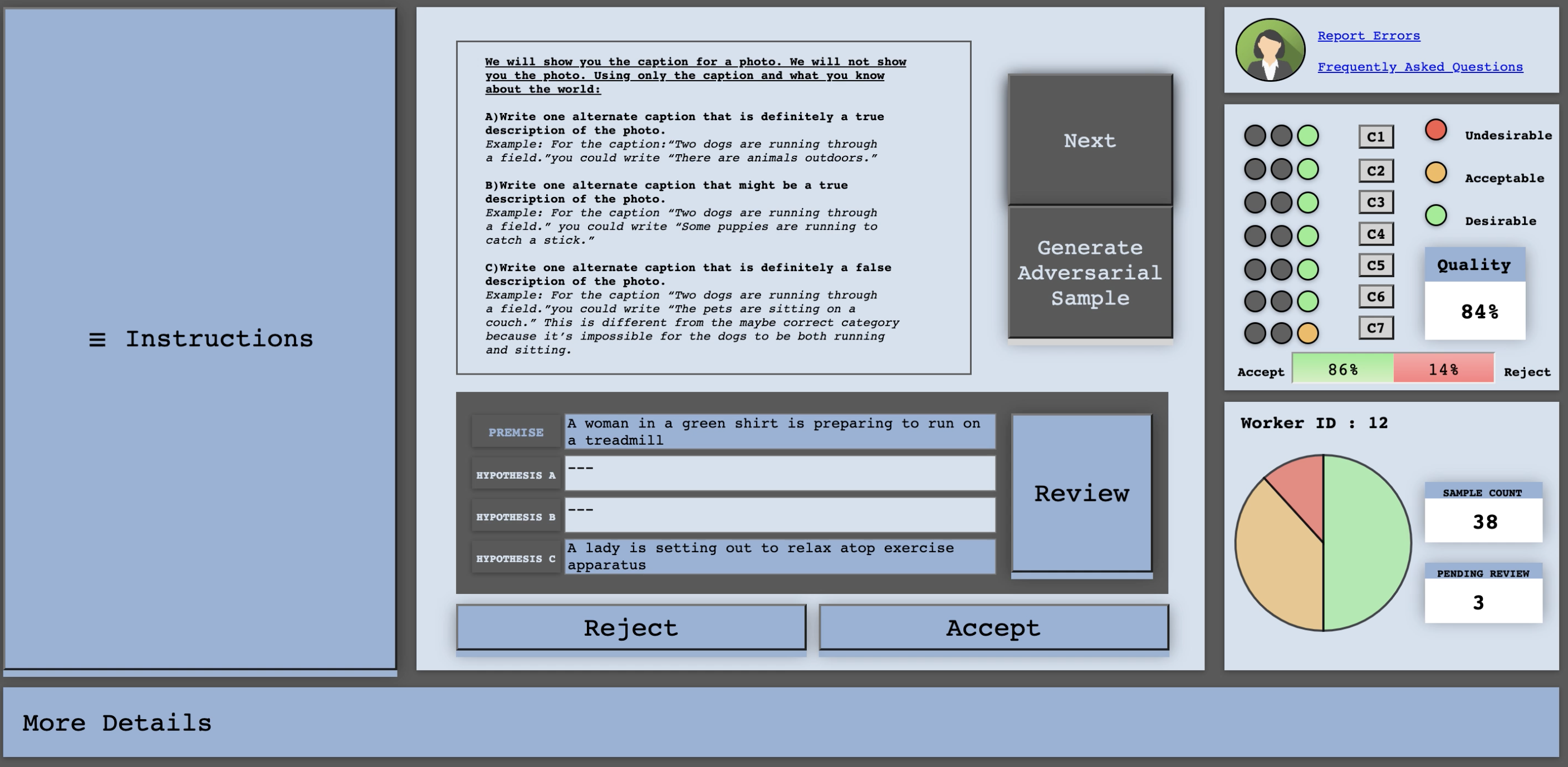}
  \caption{Analyst View}
  \label{fig:anaview}
\end{figure*}

\begin{table*}[t]
	\textsf{
      {
        \footnotesize
        \begin{tabularx}{\textwidth}{p{3cm}p{6.8cm}X}
            \toprule
            \textbf{Task} & \textbf{Description} & {Component}\\
            \midrule \rowcolor{gray!15}
            New Sample & Adds the sample under review to dataset and updates visualizations. & All\\
            Undo & Removes sample under review from dataset and updates visualizations.& All\\ \rowcolor{gray!25}
            Randomize Split & Randomized re-sampling of data across splits in a 70:10:20 ratio. & Vocabulary\\
            Undo Split & Reverses last random split generated. & Vocabulary\\ \rowcolor{gray!15}
            Save Split & Freezes split for the remainder of analysis. & Vocabulary\\
            Changing Granularity & View granularity can be changed by selecting drop down option. & Inter-sample N-gram Frequency and Relation, N-Gram Frequency per Label \\ \rowcolor{gray!15}
            Change Heat Map View & Using the drop down, the heatmap shows word similarities for the (a) premise, (b) hypothesis, or (c) both sentences. & Intra-sample Word Similarity\\
            Rebinning Histogram & By filling a new value in the textbox, the number of bins in the histogram changes to that value. & Intra-sample STS\\ \rowcolor{gray!15}
            Remove Outliers & Removes elements with frequency count less than median count of granularity being viewed.&N-Gram Frequency per Label\\
            Include All Samples & Displays all elements for a granularity.&N-Gram Frequency per Label\\
            \bottomrule
        \end{tabularx}
      }
    }
	\caption{Task Descriptions for Visual Interfaces}
	\label{tab:dqi_tasks}
\end{table*}

\begin{table*}[t]
\large
\resizebox{1.9\columnwidth}{!}{%
\begin{tabular}{llllll}
\hline
\textbf{Premise}                                                                                                    & \textbf{Orig. Hypothesis}                                                                                                                    & \textbf{DQI}                                                  & \textbf{\begin{tabular}[c]{@{}l@{}}Suggested \\ Words\end{tabular}} & \textbf{\begin{tabular}[c]{@{}l@{}}New Hypothesis based \\ on suggestions\end{tabular}}                                                              & \textbf{New DQI}  \\ \hline
\begin{tabular}[c]{@{}l@{}}A woman, in a green shirt, \\ preparing to run on a treadmill.\end{tabular}              & \begin{tabular}[c]{@{}l@{}}A woman is preparing to\\  sleep on a treadmill\end{tabular}                                                         & \begin{tabular}[c]{@{}l@{}}2.4650170\end{tabular} & preparing,sleep                                                     & \begin{tabular}[c]{@{}l@{}}A woman is organizing\\  to rest on a treadmill\end{tabular}                                                              & \begin{tabular}[c]{@{}l@{}}2.5275722\end{tabular} \\ \hline
The dog is catching a treat                                                                                         & The cat is not catching a treat                                                                                                                 & \begin{tabular}[c]{@{}l@{}}2.752542\end{tabular} & catching                                                            & the cat is not getting a treat                                                                                                                       & \begin{tabular}[c]{@{}l@{}}3.6909140\end{tabular}  \\ \hline
\begin{tabular}[c]{@{}l@{}}Three young men are watching\\ a tennis match on a large screen\\  outdoors\end{tabular} & \begin{tabular}[c]{@{}l@{}}Three young men watching \\ a tennis match on a screen \\ outdoors, because their \\ brother is playing\end{tabular} & \begin{tabular}[c]{@{}l@{}}2.6435402\\ 891414217\end{tabular} & \begin{tabular}[c]{@{}l@{}}young,watching,\\ playing\end{tabular}   & \begin{tabular}[c]{@{}l@{}}Three youthful men observing\\  a tennis match on a screen outdoors, \\ because their brother is performing.\end{tabular} & \begin{tabular}[c]{@{}l@{}}2.6787982\end{tabular} \\ \hline
\end{tabular}%
}
\caption{A few samples for Autofix with Intra Sample STS in DQI }
\label{tab:Autofixeg}
\end{table*}

\begin{table*}[t]
    \centering
    \resizebox{1.9\columnwidth}{!}{%
    \begin{tabular}{lllllll}
    \hline
         \textbf{Premise} & \textbf{Orig. Hypothesis} & \textbf{DQI} & \textbf{New Hypothesis} & \textbf{New DQI} & \textbf{Label}\\\hline
 A woman and a man sweeping the sidewalk. & The couple is sitting down for dinner. & 2.416 & The couple is meeting for dinner. & 3.479 & Contradiction \\ \hline         
A woman enjoying the breeze of a primitive fan. & The woman has a fan. & 2.127 & The woman owns a fan. & 2.733 & Entailment \\ \hline         
 There is a man in tan lounging outside in a chair. & A man is preparing for vacation. & 2.801 & A man is arranging to take a vacation. & 3.502 & Neutral \\ \hline         
    \end{tabular}%
    }
    \caption{Examples for TextFooler, with DQI's Intra-sample STS values for existing SNLI samples.}
    \label{tab:tfeg}
\end{table*}

\subsection{User Study}
\label{supp9}

\paragraph{AutoFix Suggestions:} See Tables \ref{tab:Autofix}, \ref{tab:tfeg}.

\begin{table*}[t]
\large
\resizebox{\textwidth}{!}{%
\begin{tabular}{llllll}
\hline
\textbf{Premise}                                                                                                    & \textbf{Orig. Hypothesis}                                                                                                                    & \textbf{DQI}                                                  & \textbf{\begin{tabular}[c]{@{}l@{}}Suggested \\ Words\end{tabular}} & \textbf{\begin{tabular}[c]{@{}l@{}}New Hypothesis based \\ on suggestions\end{tabular}}                                                              & \textbf{New DQI}  \\ \hline
\begin{tabular}[c]{@{}l@{}}A woman, in a green shirt, \\ preparing to run on a treadmill.\end{tabular}              & \begin{tabular}[c]{@{}l@{}}A woman is preparing to\\  sleep on a treadmill\end{tabular}                                                         & \begin{tabular}[c]{@{}l@{}}2.4650170\end{tabular} & preparing,sleep                                                     & \begin{tabular}[c]{@{}l@{}}A woman is organizing\\  to rest on a treadmill\end{tabular}                                                              & \begin{tabular}[c]{@{}l@{}}2.5275722\end{tabular} \\ \hline
The dog is catching a treat                                                                                         & The cat is not catching a treat                                                                                                                 & \begin{tabular}[c]{@{}l@{}}2.752542\end{tabular} & catching                                                            & the cat is not getting a treat                                                                                                                       & \begin{tabular}[c]{@{}l@{}}3.6909140\end{tabular}  \\ \hline
\begin{tabular}[c]{@{}l@{}}Three young men are watching\\ a tennis match on a large screen\\  outdoors\end{tabular} & \begin{tabular}[c]{@{}l@{}}Three young men watching \\ a tennis match on a screen \\ outdoors, because their \\ brother is playing\end{tabular} & \begin{tabular}[c]{@{}l@{}}2.6435402\\ 891414217\end{tabular} & \begin{tabular}[c]{@{}l@{}}young,watching,\\ playing\end{tabular}   & \begin{tabular}[c]{@{}l@{}}Three youthful men observing\\  a tennis match on a screen outdoors, \\ because their brother is performing.\end{tabular} & \begin{tabular}[c]{@{}l@{}}2.6787982\end{tabular} \\ \hline
\begin{tabular}[c]{@{}l@{}}A man in a green apron smiles\\  behind a food stand\end{tabular}                        & A man smiles                                                                                                                                    & \begin{tabular}[c]{@{}l@{}}3.2367785\end{tabular}  & smiles                                                              & A person is grinning.                                                                                                                                & \begin{tabular}[c]{@{}l@{}}6.303777\end{tabular}  \\\hline
\end{tabular}%
}
\caption{A few samples for Autofix with ISSTS in DQI }
\label{tab:Autofix}
\end{table*}

\paragraph{NASA TLX:}

The NASA Task Load Index (NASA-TLX) is a subjective, multidimensional assessment tool that rates perceived workload in order to assess a task, system, or team's effectiveness or other aspects of performance \cite{hart2006nasa}. 

\begin{figure*}[!t]
    \centering
    \includegraphics[width=0.7\textwidth]{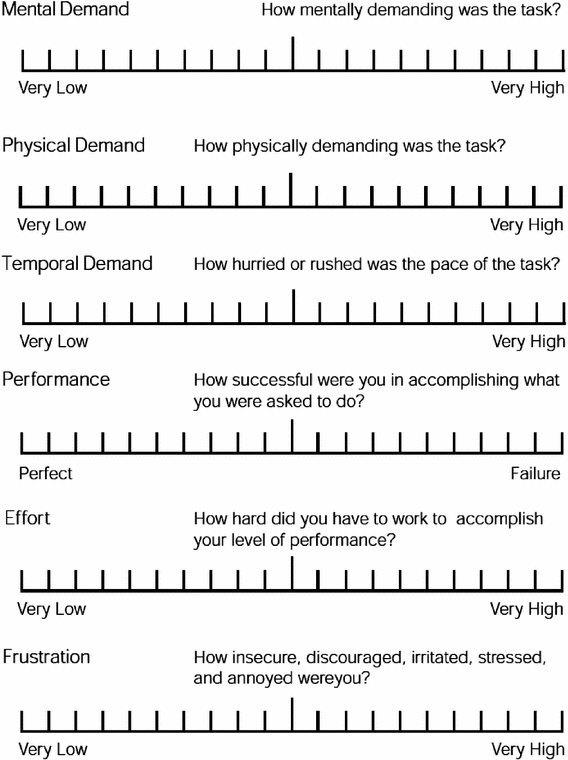}
    \caption{NASA TLX Form}
    \label{fignasa}
\end{figure*}

NASA-TLX divides the total workload into six subjective subscales that are represented on a single page. There is a description for each of these subscales that the subject should read before rating. They rate each subscale within a 100-point range, with 5-point steps, as shown in Figure \ref{fignasa}. Providing descriptions for each measurement can be found to help participants answer accurately \cite{schuff2011comparing}. The descriptions are as follows:

\begin{itemize}
    \item \textbf{Mental Demand: } How much mental and perceptual activity was required? Was the task easy or demanding, simple or complex?
    \item \textbf{Physical Demand: } How much physical activity was required? Was the task easy or demanding, slack or strenuous?
        \item \textbf{Temporal Demand: } How much time pressure did you feel due to the pace at which the tasks or task elements occurred? Was the pace slow or rapid?
            \item \textbf{Performance: } How successful were you in performing the task? How satisfied were you with your performance?
                \item \textbf{Effort: } How hard did you have to work (mentally and physically) to accomplish your level of performance?
                    \item \textbf{Frustration: } How irritated, stressed, and annoyed versus content, relaxed, and complacent did you feel during the task?
\end{itemize}

\begin{figure*}
    \centering
    \includegraphics[width=\textwidth]{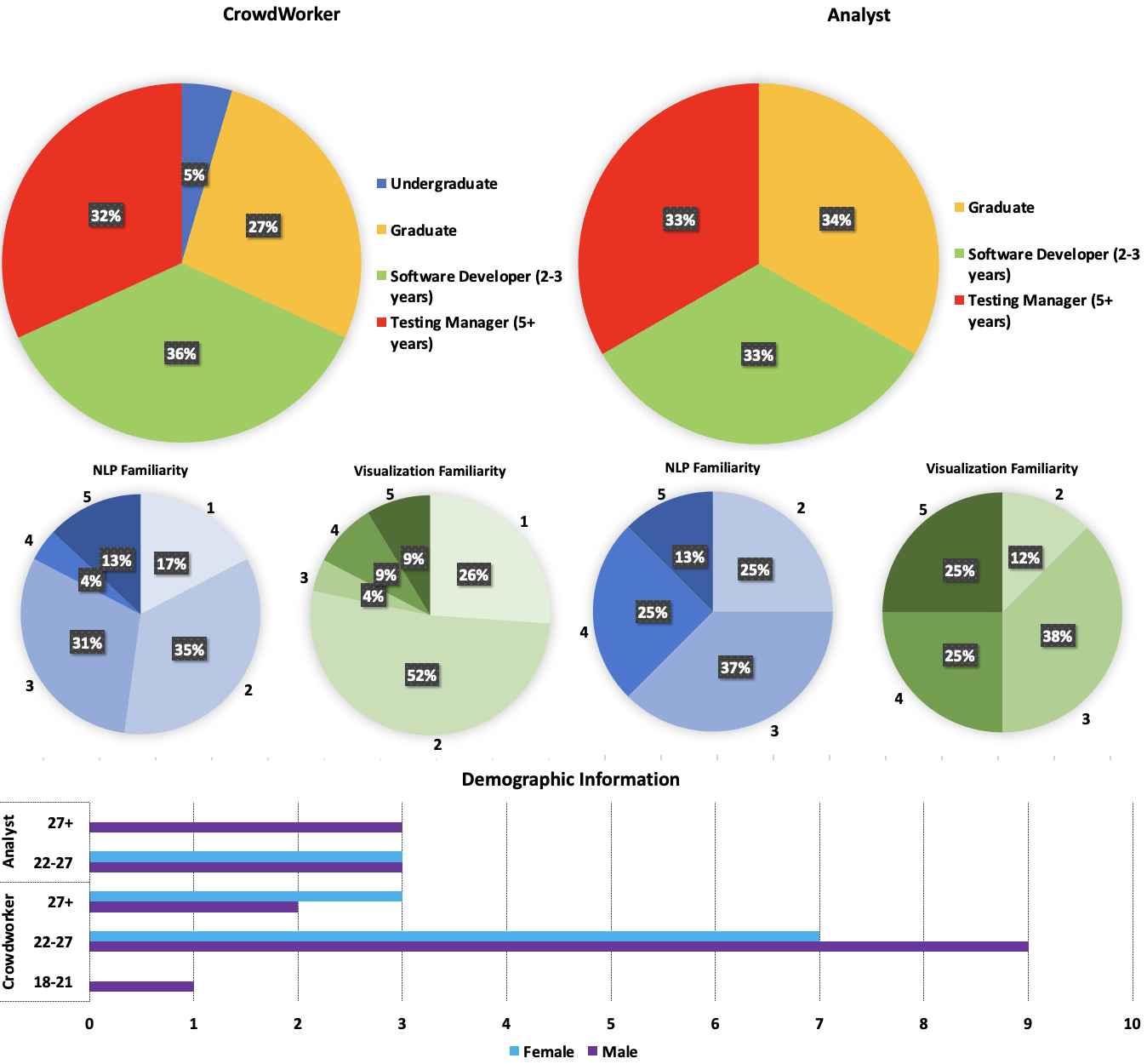}
    \caption{Demographic information for the User Study}
    \label{fig:demouser}
\end{figure*}

We record participant demographics-- age, gender, and occupation. We also ask participants to rate their familiarity with Visualization and NLP, on a scale of 1 (novice) to 5 (expert). Demographic information is shown in Figure \ref{fig:demouser}. Participants are asked to fill this form at the end of each round of the user study. We also record the number of questions participants successfully create, as well as a record of how often participants use each module in the full system round. At the end of the user study, participants are asked what their impression of data quality is, and their free response is recorded.

\paragraph{Subscale Wise Results:}

\begin{figure*}[t]
    \centering
    \includegraphics[width=0.6\textwidth]{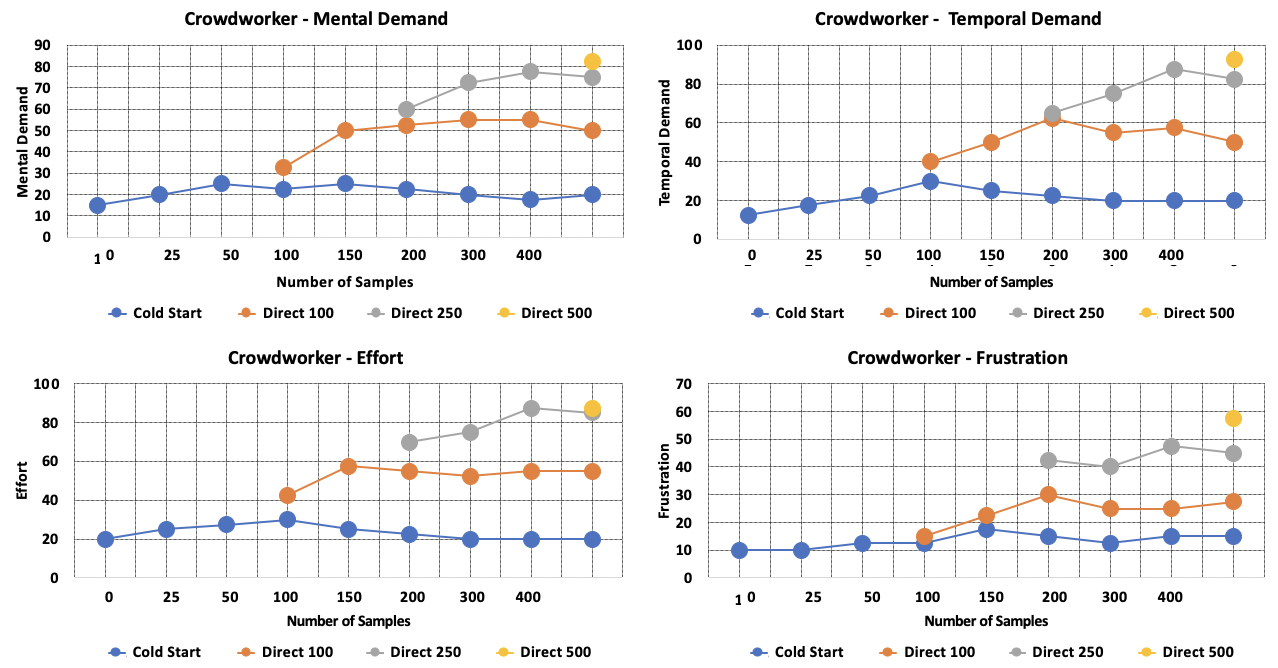}
    \caption{NASA TLX-- Crowdworker Subscale Results}
    \label{figc}
\end{figure*}

\begin{figure*}[t]
    \centering
    \includegraphics[width=0.7\textwidth]{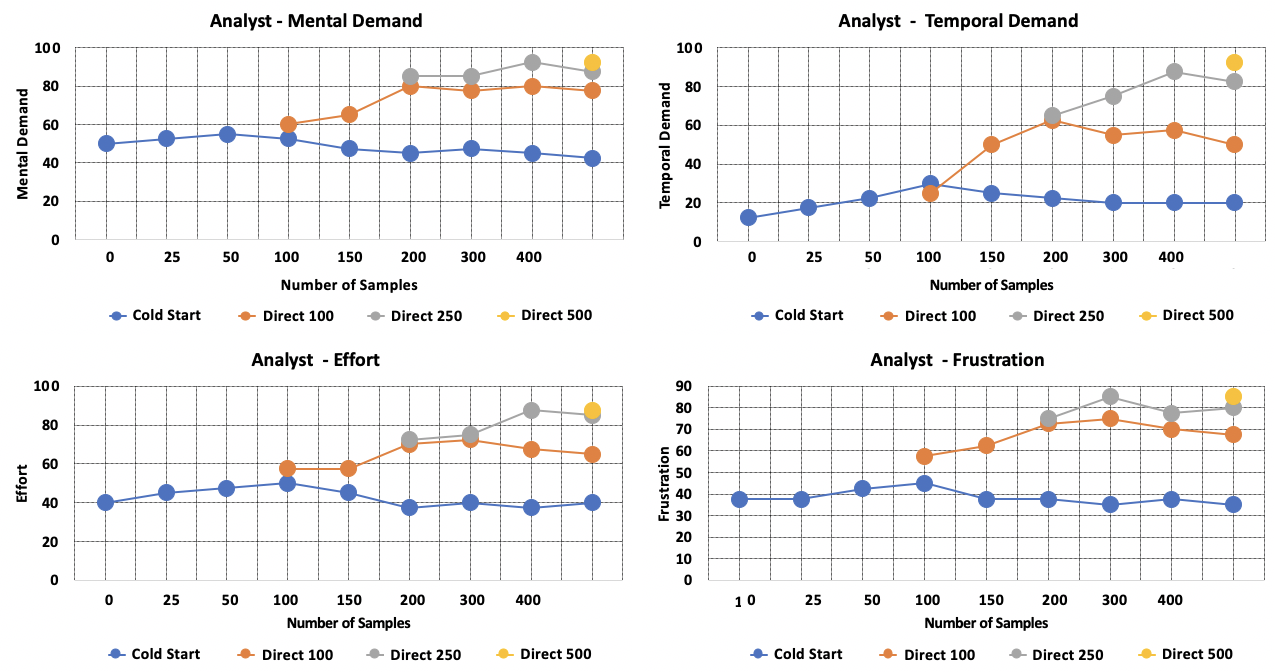}
    \caption{NASA TLX-- Analyst Subscale Results}
    \label{figa}
\end{figure*}

Individual results of the averaged subscales in Figure \ref{figuser} are shown in Figures \ref{figc},\ref{figa}. Physical demand does not change significantly across user study rounds.

\subsection{Expert and User Comments}
\label{supp10}

\textbf{Experts \textit{(P)}:} We present an initial prototype of our tool, to a set of three researchers  with expertise in NLP and knowledge of data visualization, in order to judge the interface design. For each expert, the crowdworker interface and then analyst interfaces were demoed. Participants ($P$) could ask questions and make interaction/navigation decisions to facilitate a natural user experience. All the experts appreciated the easily interpretable traffic-signal color scheme and found the organization of the interfaces---providing separate detailed views within the analyst workflow--  a way to prevent cognitive overload (too much information on one screen); \textbf{$P_2$}  said the latter \textit{``…enhances readability for understanding the data at different granularities."}.  \textbf{$P_1$} suggested the inclusion of \textit{``…a provenance module within the analyst views to show historical sample edits and overall data quality changes over time to understand how data quality evolves as the benchmark size increases… this would help with the bubble plot and tree map which will get more cluttered and complex as data size increases"}. Additionally \textbf{$P_3$} remarked that \textit{``The frequency of samples of middling quality should increase as benchmark size increases, but the initial exposure that analysts will have with higher or lower quality samples should lessen the learning curve as they are familiar enough with interface subtleties by the time they begin to encounter more challenging cases."} 

\textbf{Crowdworkers \textit{(C)}: } When presented with traffic signal feedback, crowdworkers report that the time and effort required to create high quality samples increases--\textit{``You need to keep redoing the sample since when you see it's all red, you know it's probably not going to be accepted"}\textbf{$(C_3)$}; however, they are more confident about their performance and sample quality \textit{``...when there's green, I know I've done it right, and it cuts down on my having to create a lot of samples to get paid"} \textbf{$(C_{15})$}. We find that AutoFix usage \textsuperscript{\ref{foot1}} causes an unexpected increase in mental and temporal demand, as well as frustration; we attribute this to observed user behavior-- \textit{``I'm not sure how much I trust this recommendation without seeing the colors"}\textbf{$(C_{12})$}, and \textit{``I'd prefer to change a couple of things since I can't see the feedback anymore}\textbf{$(C_{21})$}.  The drastic improvement over all aspects (highest for frustration) in the case of using the full system is in line with this observation--\textit{``This is so easy, I can create samples really fast, and I have a better chance of getting more accepted."}\textbf{$(C_{8})$} and \textit{``Now that I get the feedback along with the recommendation, I can see the quality improvement. So using the recommendation is now definitely faster."}\textbf{$(C_{12})$}. The number of questions created per round as well as system scores also follows this trend, across all types of crowdworkers.

\textbf{Summary:} Traffic signal feedback initially increases time (\textcolor{incG}{+25\%}) and effort (\textcolor{incG}{+60\%}) required to create high quality samples, as users have to correct them. However they are more confident (performance-- \textcolor{incG}{+27\%}) of sample quality. AutoFix usage causes an unexpected increase in effort (\textcolor{incG}{+5\%}) and  frustration (\textcolor{incG}{+88.8\%}), as users do not fully trust recommendations without visual feedback. The drastic improvement over all aspects (frustration-- \textcolor{decR}{-44.4\%}, mental demand-- \textcolor{decR}{-38.1\%}, temporal demand-- \textcolor{decR}{-29.1\%}, effort-- \textcolor{decR}{-20\%}, average decrease in difficulty-- \textcolor{decR}{-31.1\%}, performance-- \textcolor{incG}{+34.6\%}) in the case of using the full system is in line with this observation. The number of questions created per round (traffic signal-- \textcolor{decR}{-8.3\%}, AutoFix-- \textcolor{incG}{+25\%}, full system-- \textcolor{incG}{+83.3\%}) as well as system scores (traffic signal-- \textcolor{incG}{+27.3\%}, AutoFix-- \textcolor{incG}{+13.6\%}, full system-- \textcolor{incG}{+54.5\%}) also follows this trend, across all types of crowdworkers.

\textbf{Analysts \textit{(A)}: } In the case of direct quality feedback, i.e., traffic signals, analysts report an increased performance and find the task easier--\textit{``... it's easier to directly choose based on quality... and it takes care of typos too, the typo samples are marked down so the work goes pretty fast"}($A_3$). When analysts are shown the visualization interfaces, they are explicitly taught to differentiate the traffic signal colors in the visualizations as being indicative of how the sample affects the overall dataset quality, i.e., the colors in different component views represent individual terms of the components calculated over the whole dataset (analysts can toggle between the states of original dataset and new sample addition). We find that users initially find this more difficult to do-- \textit{``It takes a little time to figure out how to go through the views. I learned that in the samples I looked at, components three and seven seemed to be linked. So I'd look at those first the next time I used the system"} ($A_6$) and \textit{``... it takes me some time to figure out how to read the interfaces effectively, but it does make me more secure in judging sample quality at multiple granularities and that would help if I was doing this for a particular application"}($A_1$). Analysts averaged behavior on TextFooler models the conventional approach quite closely, as analysts are seen to have a tendency to either-- \textit{``... deciding to reject or repair is difficult when you don't have the sample or dataset feedback... and what if the repaired sample still isn't good enough?"}($A_4$), or--  \textit{`` I like having this option to repair... I don't need to waste time on analyzing something that isn't outright an accept or reject, I can send it to be repaired and come back to it later"}($A_8$).  When shown the full system, analysts also report improvement in all aspects, particularly mental demand and performance--\textit{``I can be sure of not having to redo things since it's likely that I will be able to get a low hypothesis baseline using this system"}($A_2$, $A_1$). The visualization usage also improves-- \textit{``... I went to component three right off the bat this time, I knew that I could look at the linked components..."} ($A_6$). Altogether, sample evaluation by analysts increases, following this trend, and analysts are more assured of their performance.

 \textbf{Summary:} Analysts find the task easier (effort-- \textcolor{decR}{-19.3\%}, performance-- \textcolor{incG}{+26.9\%}) with traffic signal feedback, as quality is clearly marked. When analysts are shown the visualization interfaces, they are explicitly taught to differentiate how the traffic signal colors in the visualizations indicate a sample's effect on the overall dataset quality. Analysts can toggle between the states of original dataset and new sample addition. We find that analysts initially find toggling more difficult to do (mental demand-- \textcolor{incG}{+15.4\%}, temporal demand-- \textcolor{incG}{+36.4\%}, frustration-- \textcolor{incG}{3.5\%}), though they agree that it improves their judgement of quality (performance-- \textcolor{incG}{+15.9\%}). Analysts' average behavior on TextFooler models the conventional approach quite closely, as analysts are seen to have a tendency to send all samples that are unclear to TextFooler immediately. With the full system, analysts also report improvement in all aspects (average decrease in difficulty-- \textcolor{decR}{-14.3\%}), particularly mental demand (\textcolor{decR}{-19.2\%}) and performance (\textcolor{incG}{+30.8\%}), considering that the system increases the likelihood of a low hypothesis baseline. The visualization usage also improves, as analysts learn component relationships. Altogether, sample evaluation by analysts increases (full system-- \textcolor{incG}{+83.3\%}), following this trend, and analysts are more assured of their performance (full system score-- \textcolor{incG}{+94.1\%}).

\end{document}